%% file: Main.tex
\documentclass[journal]{IEEEtran}
\usepackage{graphicx}
\usepackage{subfigure}  

\usepackage[square,numbers]{natbib}

\usepackage{amssymb}
\usepackage{times}
\usepackage{array}
\usepackage{soul}
\usepackage{url}
\usepackage[hidelinks]{hyperref}
\usepackage[utf8]{inputenc}
\usepackage[justification=centering]{caption}
\usepackage{graphicx}
\usepackage{amsmath}
\usepackage{booktabs}
\usepackage{algorithm}
\usepackage{algorithmic}
\usepackage{multirow}
\usepackage{amsfonts}
\usepackage{diagbox}
\usepackage{color}
\usepackage{lineno}

\newcommand{\comment}[1]{}

\begin{document}

\title{Building and Using Personal Knowledge Graph to Improve Suicidal Ideation Detection\\
on Social Media}

\author{Lei~Cao, Huijun~Zhang,
        and~Ling~Feng~\IEEEmembership{Senior~Member,~IEEE}
\IEEEcompsocitemizethanks{\IEEEcompsocthanksitem The authors are with the Department
of Computer Science and Technology,
Centre for Computational Mental Healthcare,
Research Institute of Data Science,
Tsinghua University, Beijing, China.\protect\\
E-mail: cao-l17@mails.tsinghua.edu.cn, zhang-hj17@mails.tsinghua.edu.cn,
fengling@mail.tsinghua.edu.cn}}

\maketitle

\begin{abstract}
A large number of individuals are suffering from suicidal ideation in the world.
There are a number of causes behind why an individual might suffer from suicidal ideation.
As the most popular platform for self-expression, emotion release, and personal interaction,
individuals may exhibit a number of symptoms of suicidal ideation on social media.
Nevertheless, challenges from both data and knowledge aspects
remain as obstacles, constraining the social media-based detection performance.
Data implicitness and sparsity make it difficult to discover the inner true intentions of individuals
based on their posts.
Inspired by psychological studies, we build and unify a high-level suicide-oriented knowledge graph
with deep neural networks for suicidal ideation detection on social media.
We further design a two-layered attention mechanism to
explicitly reason and establish key risk factors to individual's suicidal ideation.
The performance study on microblog and Reddit shows that: 1) with the constructed personal knowledge graph,
the social media-based suicidal ideation detection can achieve over 93\% accuracy;
and 2) among the six categories of personal factors, \emph{post, personality,} and \emph{experience} are the
top-3 key indicators. Under these categories, \emph{posted text}, \emph{stress level},
\emph{stress duration}, \emph{posted image}, and \emph{ruminant thinking} contribute to one's suicidal ideation detection.
\end{abstract}

\begin{IEEEkeywords}
Suicidal ideation detection, social media, personal knowledge graph, social interaction.
\end{IEEEkeywords}

\IEEEpeerreviewmaketitle

\input{new-Sec1-Introduction}

\input{new-Sec2-Relatedwork}
\input{new-Sec3-Methods-KG-new}

\input{new-Sec3-Methods-GNN}

\input{new-Sec4-Results-new}

\input{new-Sec5-Discussion}

\input{new-Sec6-Conclusion}

\section*{Acknowledgment}
We thank all the anonymous reviewers' constructive comments, enabling us to improve the manuscript.

The work is supported by the National Natural Science Foundation of China (61872214, 61532015, 61521002).

\footnotesize{
\bibliographystyle{IEEEtran}
\bibliography{suicide}}

\begin{IEEEbiography}[{\includegraphics[width=1.1in,height=1.4in,clip,keepaspectratio]{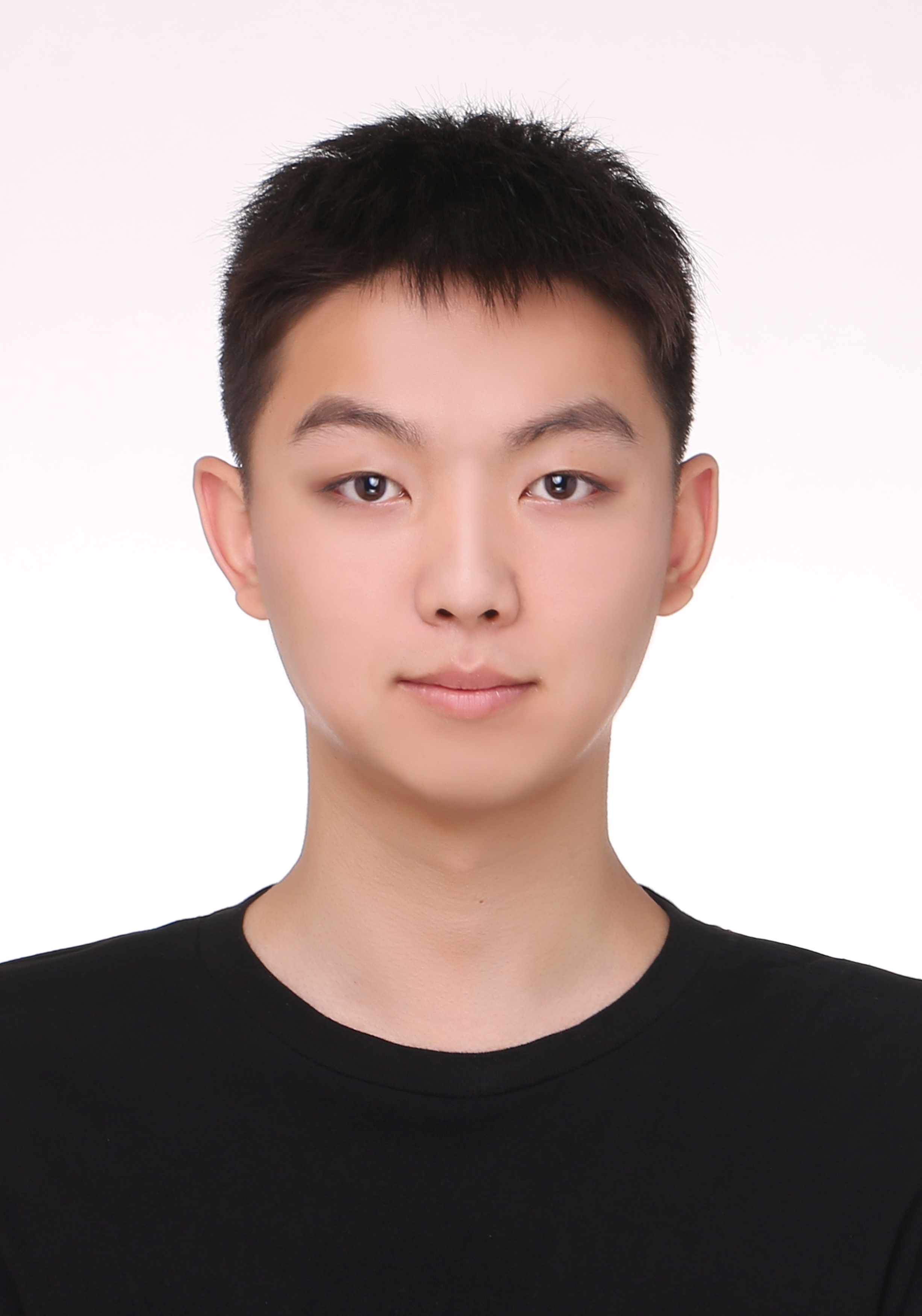}}]{Lei Cao}
is a PhD candidate in the Department of Computer Science and Technology, Tsinghua University, Beijing, China.
His research interests include computational psychology and sentiment analysis.
\end{IEEEbiography}
\begin{IEEEbiography}[{\includegraphics[width=1in,height=1.25in,clip,keepaspectratio]{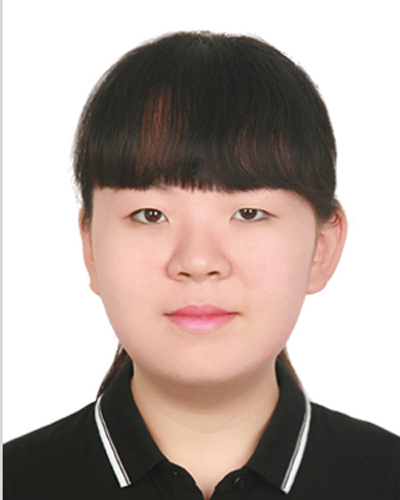}}]{Huijun Zhang} is a PhD candidate in the Department of Computer Science and Technology, Tsinghua University, Beijing, China.
Her research interests include computational psychology and sentiment analysis.
\end{IEEEbiography}
\begin{IEEEbiography}[{\includegraphics[width=1in,height=1.25in,clip,keepaspectratio]{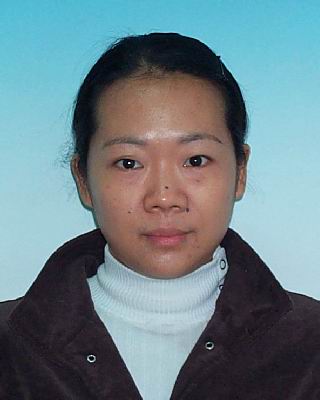}}]{Ling Feng}
is a professor of computer science and technology with Tsinghua University, China.
Her research interests include computational mental healthcare, context-aware data management and services toward ambient intelligence, data mining and warehousing, and distributed object-oriented database systems.
\end{IEEEbiography}

\end{document}

%% file: new-Sec1-Introduction.tex
\section{Introduction}

\IEEEPARstart{S}{uicidal} ideation is a problem which is on the rise across all the nations in the world.
According to different reports and statistical figures, the lifetime prevalence
of suicidal ideation of the entire world population is around
9 per cent~\cite{Nock2008}, and
this figure is significantly higher for individuals in the age group of 18 to 25 years old.
There are a number of causes behind why an individual might suffer from suicidal ideation, and
involvement of personality traits in susceptibility to suicide has
been investigated for a long time. However, due to the limitations of mass group reaching and
the diversity of conceptual and methodological approaches,
the extent of their independent contributions remain is still have been difficult to establish.
The risk factors for the immediate precursor to suicidal ideation are still
not well-known especially in developing countries~\cite{Nock2008}.
In any circumstance, for those with suicidal ideation, the earlier they can be detected,
the better the chance of suicide prevention~\cite{Jacob2014}.


In the literature, psychologists have developed a number of
suicide risk measurements (such as
Suicide Probability Scale~\cite{bagge1998suicide},
Adult Suicide Ideation Questionnaire~\cite{fu2007predictive},
Suicidal Affect-Behavior-Cognition Scale~\cite{harris2015abc}, etc.)
to assess individual's suicidal ideation.
As this kind of methods requires people to either fill in a
subjective questionnaire or participate in a professional interview,
it is only applicable to a small group of people.
For those who are suffering but tend
to hide inmost thoughts and refuse to seek helps from others, the
approach cannot work~\cite{E-health2014,essau2005frequency,rickwood2007and,zachrisson2006utilization}.

%

With social media (like Twitter, online forums, and microblogs)
becoming an integral part of daily lives nowadays,
more and more people go to social media for information
acquisition, self-expression, emotion release, and personal
interaction.
The large-scale, low-cost, and open advantages of social media
offer us the unprecedented opportunity to capture individual's symptoms and traits
related to suicidal ideation~\cite{ji2020suicidalb,Alambo2019,Cheng2017Assessing,Du2018,SawhneyMMSS18,coppersmith2018natural,Vioul2018Detection}.
Nevertheless, analysis and detection of individual's suicidal ideation through
social media are not trivial, facing a number of challenges from the following
two aspects.


%

\subsection{Data-Aspect Challenges}

Data implicitness and data sparsity are two critical data-aspect problems,
challenging social media-based solutions for a long time.

\textbf{Data Implicitness.}
Due to the unique equality, freedom, fragmentation,
and individuality characteristics of social media,
users linguistic and visual expressions on social media are implicit, reserved,
and even anti-real.
To illustrate, let's compare users' normal posts with their
commenting posts in a hidden \emph{microblog tree hole}.
Here, a microblog tree hole refers to
a microblog space, whose author has committed suicide, and other microblog users
usually with suicidal ideation tend to share and
post their inner real feelings and thoughts as comments
under the last post of the passed one.
One such a tree hole on Sina Weibo (the largest microblog platform in China) has collected over 1,700,000 commenting posts from
more than 160,000 suicide attempts under the last post of microblog user called
\emph{Zoufan}, who committed suicide due to depression on March 17, 2012.

\begin{figure*}[htb]
\centering
\includegraphics[scale=0.8]{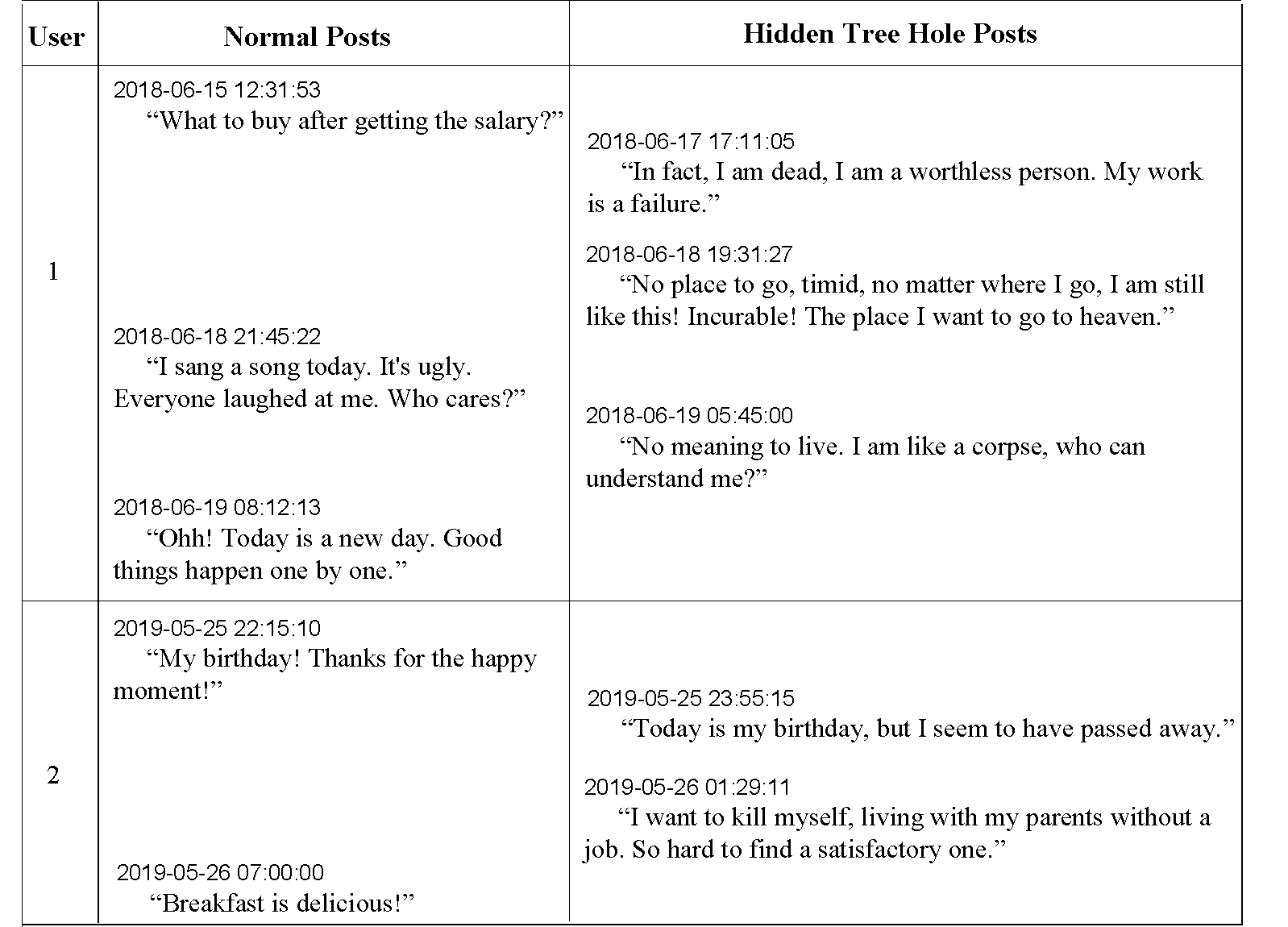}
\caption{Two users' normal posts versus their hidden tree hole posts on Sina Weibo.}
\label{fig:tab:normalHiddenposts}
\end{figure*}

Figure~\ref{fig:tab:normalHiddenposts} lists two users' example posts on microblog.
From their normal posts, we can sense a joyful feeling.
However, if reading their commenting posts in the hidden tree hole,
we are touched by their severe hopelessness and suicidal thoughts.


To further examine the differences between users' normal posts and
their commenting posts in the hidden tree hole, we crawled and analyzed  
3,652 individuals' posts from May 1, 2018 to April 30, 2019 on Sina Weibo.
As Table~\ref{tab:postContentCompare} shows,
their posting contents were quite different.
The hidden tree hole posts focused more on oneself through more self-concern
words (like ``I", ``me", ``my", ``mine", ``am", etc.) rather than others-concern words,
and they used more suicide-related words than the normal posts,
demonstrating users' willingness and directives to express their inner
suicidal thoughts in the hidden tree hole.
On the contrary, the users were reluctant to show their feelings
in the open normal posts, and they used more others-concern (such as ``they", ``their", ``she", ``he",  etc.) words than the tree hole posts.
Such phenomena challenge the suicide risk detection through the normal posts.

\begin{table}[htp]
\caption{\centering Statistics of users' normal posts vs. hidden tree hole posts on Sina Weibo based on 3,652 users from May 1, 2018 to April 30, 2019.}
\label{tab:postContentCompare}
\centering
\begin{tabular}{p{5.4cm} m{0.6cm}<{\centering} m{1.6cm}<{\centering}}
\hline\noalign{\smallskip}
\textbf{Perspective} &  \textbf{Normal} & \textbf{Hidden Tree} \\
                     &  \textbf{~Posts} & \textbf{Hole Posts} \\
\noalign{\smallskip}\hline\noalign{\smallskip}
Prop. of posts containing \emph{self-concern} words &  43\%    & 50\%   \\
Prop. of posts containing \emph{others-concern} words &  8\%    & 6\%   \\
Prop. of posts containing \emph{suicide-related} words &  31\%    & 95\%   \\
Prop. of \emph{self-concern} words per post  &  4\%    & 9\%   \\
Prop. of \emph{others-concern} words per post &  8\%    & 1\%   \\
Prop. of \emph{suicide-related} words per post &  0.02\%    & 0.3\%   \\
Total post number  &  252,901   & 190,087 \\
 \hline\noalign{\smallskip}
\end{tabular}
\end{table}

\textbf{Data Sparsity.}
Apart from data implicitness, subjective to habits, characters, emotions, etc.,
some people may not want to express themselves actively on social media.
Particularly, when immersed in hopelessness, loneliness, and suicidal thoughts,
people tend to be self-enclosed and have very low motivations to
write or leave something on social media.


\begin{figure}[htb]
\centering
\begin{tabular}{c}
\includegraphics[width=0.7\columnwidth]{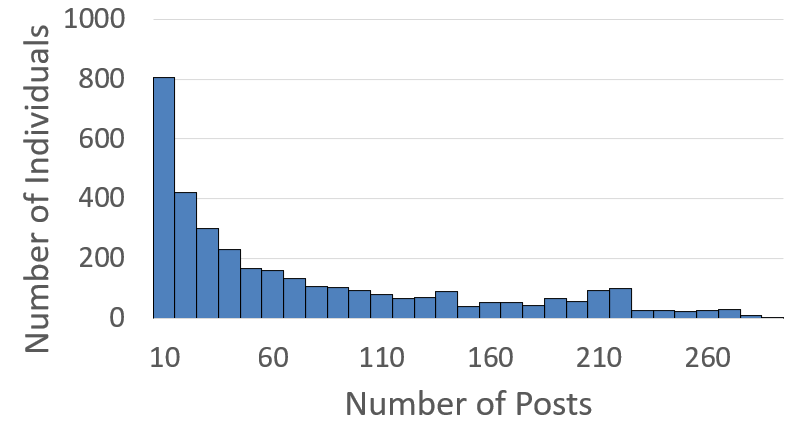}  \\
(a) Users with suicidal ideation  \\ \\ \\
\includegraphics[width=0.7\columnwidth]{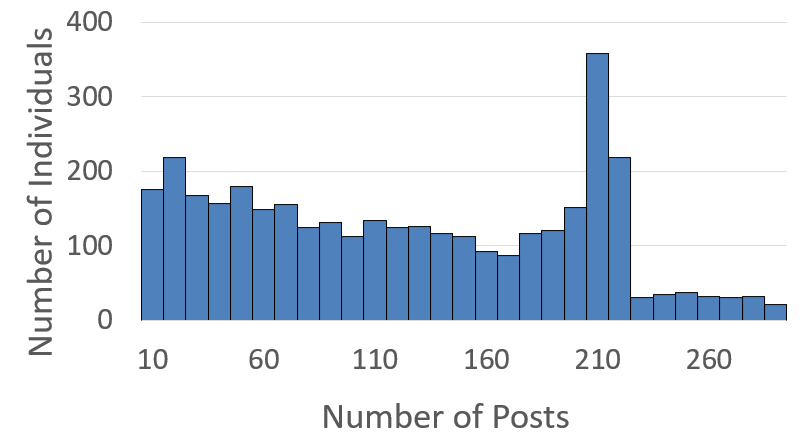}  \\
 (b) Ordinary users
\end{tabular}
\caption{Distributions of normal post numbers from 7,329 users (3,652 with suicidal ideation
and 3,677 ordinary users) from May 1,2018 to April 30,2019 on Sina Weibo.}
\label{fig:postnumComparison}
\end{figure}

\begin{table*}
\caption{Statistics of deleted normal post numbers based on 7,329 users (3,652 with suicidal ideation and 3,677 normal users) from May 1, 2018 to April 30, 2019 on Sina Weibo.}
\label{tab:datadelete}
\centering
\begin{tabular}{p{9cm} m{4.2cm}<{\centering} m{3.5cm}<{\centering}}
\hline\noalign{\smallskip}
 & \textbf{\# Users with Suicidal Ideation} & \textbf{\# Ordinary Users}  \\ 
\noalign{\smallskip}\hline\noalign{\smallskip}
Number of users who deleted their previous normal posts after 3 months & 1,673 & 326 \\ 
Proportion of users who deleted their previous normal posts after 3 months & 45.81\% & 8.87\% \\ 
Proportion of deleted normal posts after 3 months & 35.70\% & 3.95\% \\ 
\hline\noalign{\smallskip}
\end{tabular}%
\end{table*}

To illustrate, we identified 3,652 users which made both normal posts and
hidden tree hole posts on Sina Weibo, and their tree hole posts contained
suicide-related words. We hypothesized that this group of users have suicidal ideation.
For comparison purpose, we also randomly crawled another 3,677 active users,
who only posted normal blogs on microblog, and their blogs contained no suicide-related words at all.
We called this group of users \emph{ordinary users}.
From Figure~\ref{fig:postnumComparison}, we can find that
users with suicidal thoughts made much less normal posts than ordinary users.
Overall, the majority of this group of users had less than 1 normal post per week,
verifying the serious data sparsity problem.

Moreover, due to violent emotional fluctuations,
individuals with suicidal ideation tend to
delete their previous posts related to suicidal
ideation, striving to hide the true inner intentions.
We counted the number of normal posts from the
7,329 users (3,652 with suicidal ideation and 3,677 ordinary users)
in one year from 2018.05.01 to 2019.04.30
twice (one on 2019.04.30, and the other on 2019.08.31), and discovered that
(1) around 45.81\% of the users with suicidal ideation deleted their previous normal posts after 3 months,
while only 8.87\% of ordinary users did it.
(2) around 35.70\% of normal posts were deleted by
the users with suicidal ideation.
For ordinary users, the figure was just about 3.95\%.
In other words, individuals with suicidal feelings tend to delete much more normal posts than
ordinary ones (Table~\ref{tab:datadelete}).

The above data-aspect challenges hinder the detection performance greatly.

\begin{figure}
\centering
\includegraphics[scale=0.38]{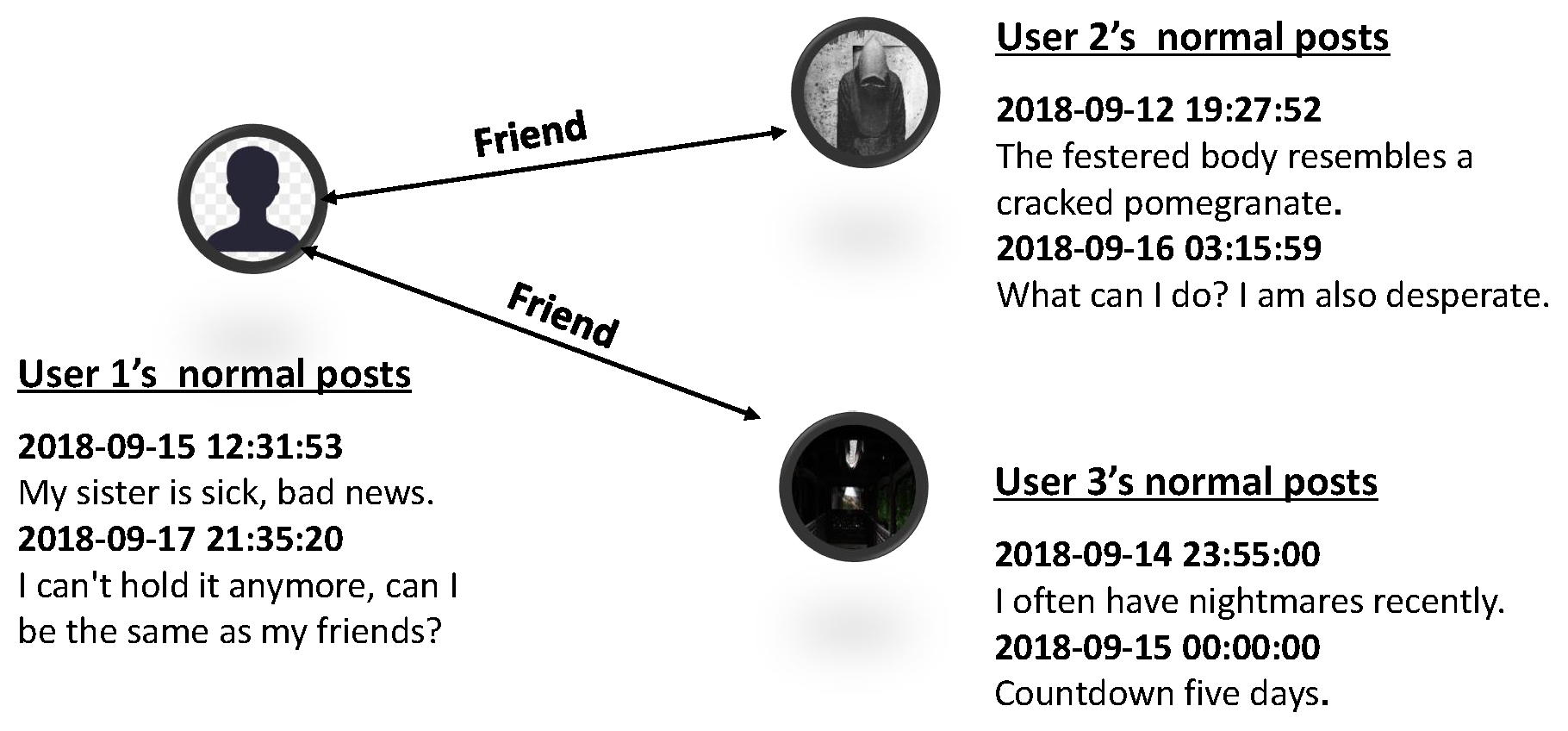}
\caption{Examples of normal posts by user 1 and his two followed users on Sina Weibo.}
\label{fig:followerpost}
\end{figure}

\textbf{Our Work. }
To address the data-aspect challenges,
we drew inspirations from previous psychological studies that
individuals tend to find and follow one's own kind in the networks,
looking for similar people, comfort each other, find a way to get rid of the difficulties, or even commit suicide
together~\cite{Robinson2015}.
In such an emotional sharing and resonance process, negative
suicidal emotion and influence could be developed and fast propagated among potentially
suicidal individuals~\cite{Madelyn2003}.
As in Figure~\ref{fig:followerpost}, we can notice similar negative feelings
from user 1 and his two followed friends.
In our crawled 7,329 users, each user follows 5 neighbour users (as friends) on average.

Motivated by the findings, we leveraged individual's friends information via the
follower-following relationship on microblogs to
enrich users data and
cope with data implicitness and sparsity problems in
suicidal ideation detection.

\subsection{Knowledge-Aspect Challenges}

Although there has been a large body of research
addressing the task of suicidal ideation based on social media,
the detection performance remains to be constrained due to
the disparity between the informal language used
by social media users and the concepts defined by domain experts in medical knowledge bases~\cite{gaur2019knowledge,Brisset2014,Nie2015}.
Moreover, methods based on domain knowledge have been successful in many fields \cite{ning2017knowledge,xue2019knowledge,chaudhary2019enhancing}.
To fill the gap, recently, \cite{gaur2019knowledge} incorporated domain specific knowledge
into a learning framework to predict the severity of
suicide risk for a Redditor user.
Specifically, it employed two medical lexicons
(TwADR-L and AskaPatient) to map social media contents to
medical concepts~\cite{limsopatham-collier-2016-normalising},
and identified contents with negative
emotions based on anonymized and annotated suicide
notes available from Informatics for Integrating Biology
and the Bedside (i2b2) challenge.
It further developed a suicide risk severity lexicon using medical knowledge bases and
suicide ontology to detect cues relevant to suicidal thoughts and
actions, and used language modeling, medical entity recognition
and normalization and negation detection to create a gold standard dataset of
500 Redditors developed by four practicing psychiatrists~\cite{gaur2019knowledge}.

The work~\cite{gaur2019knowledge} proved that utilizing domain specific knowledge is a
promising approach for suicidal ideation analysis.
In reality, there could be a large number of causes behind why an individual might suffer from suicidal ideation.
It is of extreme importance that the suicidal ideation detection method
should be able to recognize the warning signs or symptoms around the individual who is suffering,
and involve them to improve the detection performance.
However, few contribution measurements of different indicators in relation to individual's
\emph{personal information} (display image on social media, gender, age, etc.),
\emph{personality} (pursuing perfection, ruminant thinking, or interpersonal sensitivity),
\emph{experience} (stress or previous suicide attempt),
negative emotion and emotion transition pattern, and social interaction impact
have been established for suicidal ideation detection.
In the literature, so far there has no social media-based \emph{personal} suicide-oriented
knowledge framework
being developed and incorporated into suicidal ideation analysis.




\textbf{Our Work. }
Beyond looking at individual's series of posting contents and social
interactions on social media~\cite{Pradyumna2019a,Profile2019},
we built a structured personal suicide-oriented knowledge graph,
and instantiated the personal knowledge graph
by applying deep learning techniques~\cite{DBLP:conf/iclr/VelickovicCCRLB18}
to 7,329 microblog users (3,652 with suicidal ideation and 3,677 without suicidal ideation)
and 500 Redditors (108 without suicidal ideation, 99 with suicide indicator, 171 with suicidal ideation, 77 with suicide behavior and 45 with actual attempt).

We further built a two-layered attention mechanism
(property attention and neighbour attention)
to propagate model messages through the designed personal knowledge graph,
and adaptively tune the weights (contributions) of
different risk factors (properties and
neighbour users) to individuals' suicidal ideation.

Finally, we incorporated and unified the achieved personal suicide-oriented knowledge graph
with graph neural networks to facilitate
suicidal ideation detection. Our experimental results on microblog and Reddit
showed that with the personal suicide-oriented knowledge graph and the two attention mechanisms,
suicidal ideation detection can achieve over 93\% of accuracy and F1-measure.
Among the six categories of personal factors, \emph{posts, personality,} and \emph{experience} are the
top-3 key indicators. Under these categories, \emph{posted texts}, \emph{stress level},
\emph{stress duration}, \emph{posted image}, and \emph{ruminant thinking} contribute the
most to one's suicidal ideation detection.

In summary, the contributions of the study are three-fold.
\begin{itemize}
\item We build and unify a high-level suicide-oriented knowledge graph
with graph neural networks for suicidal ideation detection on social media.

\item We design a two-layered attention mechanism that explicitly reasons
and establishes key risk factors to individual's suicidal ideation.

\item We deliver a social media-based suicide-oriented knowledge graph. Together with
the labeled dataset ({\url{https://github.com/bryant03/Sina-Weibo-Dataset}}) with 3,652 (3,677) users with (without) suicidal ideation from Sina Weibo,
they could facilitate further suicide-related
studies in the computer science and psychology fields.
\end{itemize}

%% file: new-Sec2-Relatedwork.tex
\section{Literature Review on Detecting Suicidal Ideation}

\subsection{Traditional Questionnaire-based Suicide Risk Assessment}

Traditional methods rely on questionnaires and face-to-face diagnosis to assess whether one
is in the risk of suicide~\cite{pestian2017machine}.
Some widely used psychological measurements include
Suicide Probability Scale (SPS)~\cite{bagge1998suicide}, Depression Anxiety Stress Scales-21 (DASS-21)~\cite{crawford2003depression,henry2005short},
Adult Suicide Ideation Questionnaire~\cite{fu2007predictive},
Suicidal Affect-Behavior-Cognition Scale~\cite{harris2015abc},
functional Magnetic Resonance Imaging (fMRI) signatures~\cite{just2017machine}, and so on.
While this kind of approaches are professional and effective,
they require respondents to either fill in a questionnaire or participate in a professional interview,
constraining its touching to suicidal people who have low motivations to seek help from professionals~\cite{E-health2014,essau2005frequency,rickwood2007and,zachrisson2006utilization}.
A recent study found out that taking a suicide assessment may bring negative effect to
individuals with depressive symptoms~\cite{harris2017suicide}.

\subsection{Suicide Risk Detection from Social Media}

Recently, detection of suicide risk from social media is making great progress
due to the advantages of reaching massive population, low-cost, and real-time~\cite{braithwaite2016validating}.
\cite{harris2013} reported that suicidal users tend to spend a lot of time online,
have great likelihood of developing online personal relationships, and great use of online forums.

\textbf{Suicide Risk Detection from Suicide Notes.}
\cite{JP10} built a suicide note classifier used machine learning techniques, which performs
better than human psychologists in distinguishing fake online suicide notes from real ones.
\cite{YH07} hunted suicide notes based on lexicon-based keyword matching on MySpace.com (a popular site for adolescents and young adults, particularly sexual minority adolescents with over 1 billion registered users worldwide) to check whether users have an intent to commit suicide.

\textbf{Suicide Risk Detection from Community Forums.}
\cite{MT13} applied textual sentiment analysis and summarization techniques to users' posts and posts' comments in a Chinese web forum in order to identify suicide expressions.
\cite{NM14} examined online forums in Japan, and discovered that
the number of communities which a user belongs to, the intransitivity,
and the fraction of suicidal neighbors in the social network
contributed the most to suicidal ideation.
\cite{MC16} built a logistic regression framework to
analyze Reddit users' shift tendency from mental health sub-communities to a suicide support sub-community.
heightened self-attentional focus, poor linguistic coherence and coordination with the community, reduced social engagement and manifestation of hopelessness, anxiety, impulsiveness and loneliness in shared contents
are distinct markers characterizing these shifts.
Based on the suicide lexicons detailing
suicide indicator, suicidal ideation, suicide behavior, and suicide attempt,
\cite{Alambo2019} built four corresponding semantic
clusters to group semantically similar posts on Reddit and questions in a questionnaire together, and
used the clusters to assess the aggregate suicide risk severity of a Reddit post.
\cite{gaur2019knowledge} used Reddit as the unobtrusive data source
to conduct knowledge-aware assessment of severity of suicide risk for
early intervention. Its performance study showed that convolutional neural
network (CNN) provided the best performance due
to the discriminative features and use of domain-specific knowledge
resources, in comparison to SVM-L that has been used in the
state-of-the-art tools over similar datasets.
\cite{shing2020prioritization} introduced a new evaluation measure to move beyond suicide risk classification to a paradigm in which prioritization is the focus.
The proposed joint ranking approach outperformed the logistic regression under the new evaluation measure.
\cite{jones2019analysis} used document embeddings generated by transfer learning to classify users at suicide risk or not.
\cite{tadesse2020detection} employed an LSTM-CNN combined model to evaluate and compare to other classification models. 

\textbf{Suicide Risk Detection from Blogs.}
\cite{JJ14} used search keywords and phrases relevant to suicide risk factors
to filter potential suicide-related tweets,
and observed a strong correlation between
Twitter-derived suicide data and real suicide data, showing that
Twitter can be viewed as
a viable tool for real-time monitoring of suicide risk factors on a large scale.
The correlation study between suicide-related tweets and suicidal behaviors was also conducted
based on a cross-sectional survey~\cite{HS15},
where participants answered a self-administered online questionnaire, containing questions about Twitter use, suicidal behaviour, depression and anxiety, and demographic characteristics.
The survey result showed that Twitter logs could help identify suicidal young Internet users.

Based on eight basic emotion categories
(joy, love, expectation, anxiety, sorrow, anger, hate, and surprise),
\cite{FR16} examined three accumulated emotional traits (i.e., emotion accumulation, emotion covariance,
and emotion transition) as the special statistics of emotions
expressions in blog streams for suicide risk detection.
A linear regression algorithm based on the three accumulated emotional traits was employed
to examine the relationship between emotional traits and suicide risk.
The experimental result showed that by combining all of three emotion traits together,
the proposed model could generate more discriminative suicidal prediction performance.

Natural language processing and machine learning techniques, such as
Latent Dirichlet Allocation (LDA), Logistic Regression, Random Forest,
Support Vector Machine, Naive Bayes, Decision Tree, etc., were applied to
identify users' suicidal ideation
based on their linguistic contents and online behaviors on Sina Weibo~\cite{LG14,LZ15,huang2015topic,guan2015identifying,Cheng2017Assessing}
and Twitter~\cite{o2015detecting,coppersmith2015quantifying,parraga2019unsupervised,fodeh2019using,ji2020suicidal,DBLP:journals/phat/MalhotraJ20,DBLP:journals/ijkbo/Bouarara20}.
Deep learning based architectures like
Convolutional Neural Network (CNN), Recurrent Neural Network (RNN), Long Short-Term Memory Neural Network (LSTM), etc.,
were also exploited to detect users' suicide risk on social media~\cite{Du2018,SawhneyMMSS18,Sawhney2018,coppersmith2018natural}.
\cite{Vioul2018Detection} detected users' change points in emotional well-being
on Twitter through a martingale framework, which is widely
used for change detection in data streams.

Very recently, \cite{lei2019} designed suicide-oriented work embeddings to capture the implicit suicidal expression and proposed a suicide risk detection model based on LSTM and ResNet.
\cite{Profile2019} applied the machine learning algorithms to detect suicidal profiles in Twitter
based on features extracted from users accounts and tweeting behaviors.
\cite{Pradyumna2019a,Pradyumna2019b} proposed a multipronged
approach and implemented different neural network models such as
sequential models and graph convolutional networks, which were trained
on textual content shared in Twitter, tweeting histories of the users
and social network formed between different users posting about suicidality.


Different from the previous work,
in this study, we looked beyond individual's posting contents and
social interactions on social media, and
incorporated the constructed personal knowledge graph into
suicidal ideation detection. The performance study showed
that the proposed method
can achieve over 93\% of accuracy and F1-measure.

%% file: new-Sec3-Methods-KG-new.tex
\section{Framework}

\subsection{Building the Personal Suicide-Oriented Knowledge Graph}

\begin{figure}[htp]
\centering
\includegraphics[scale=0.53]{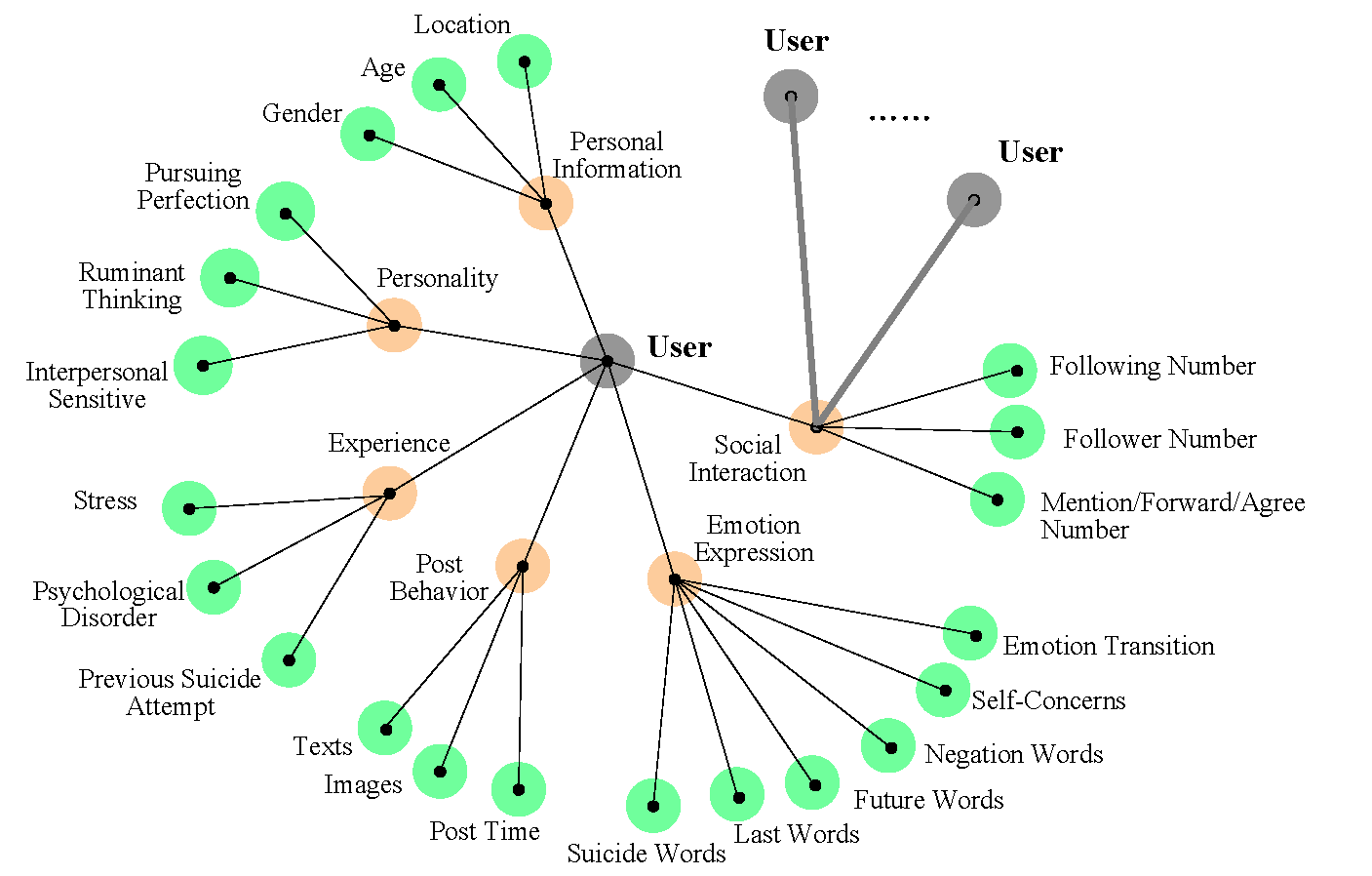}
\caption{The ontology of the social media-based suicide-oriented knowledge graph.}
\label{fig:ontology}
\end{figure}

We define the terminology for representing the nodes and edges of the
personal suicide-oriented knowledge graph, and
then convert the extracted data from individuals' social media accounts
to instantiate such a personal knowledge graph.
Inspired by the psychological investigation into the predictors of suicide risk \cite{cooperman2005suicidal,self-affirmation1988,Rich1987,bluml2013personality,greenspon2014there,ML15,LM14},
we extracted and analyzed individuals' relevant social media behaviors
from the following six perspectives (i.e., \emph{personal information,
personality, experience, post behavior, emotion expression,} and \emph{social interaction}),
as shown in Figure~\ref{fig:ontology}.

\subsubsection{Personal Information (Gender, Age, Location)}

Based on the previous study result that suffering women are
three times more likely than suffering men to attempt suicide~\cite{cooperman2005suicidal},
we captured user $u$'s personal details including gender, age, and location from his microblog account,
and involved them into the personal knowledge graph.



We used a 3-dimensional vector to describe user's gender information.
$Gender(u)$ = (1,0,0), (0,1,0), or (0,0,1),
representing
\emph{female}, \emph{male}, or \emph{unknown}, respectively.

We used a scalar from zero to one to represent user's age.
The scalar is obtained by dividing the user's age by the maximal user age
in the collected dataset.
$Age(u)$=$\frac{u.age}{max\_user\_age}$, where $max\_user\_age$=65.

Like user's gender expression, we used a vector of length 8 to represent user $u$'s location.
$Loc(u)$ = $(1,0,0,\cdots,0)$, $(0,1,0,\cdots,0)$,
$(0,0,1,\cdots,0)$, $\cdots$, or $(0,0,0,\cdots,1)$.
Each element of the vector represents a certain geographic direction.
It takes value 1 if the user is in the corresponding location, and 0 otherwise.
Totally 8 location directions (i.e., \emph{east, south, north, south-west, north-west, middle, north-east},
plus an \emph{unknown} location) were considered.



\subsubsection{Experience (Stress, Psychological Disorder, Previous Suicide Attempt)}

Suicidal ideation often occurs when an individual is no longer able to cope with some kind of difficult or
overwhelming situation~\cite{self-affirmation1988}.
Some of the commonly agreed causes of suicidal ideation include situations where an individual
experiences continuous stressful periods of
high stress levels and different stress categories~\cite{Rich1987}.
To detect user's stress (stress periods, stress levels, and stress categories) within the year,
we applied the algorithm~\cite{LiQiJournal} to user's posting behaviors, and captured
a series of stressful periods ${\cal S}(u)$,
each of which is of the form
$s$=$(s_p, s_l, s_c)$, where
$s \in {\cal S}(u)$, $s_p$ is a temporal stress period which should be over five days,
$s_l$ is the stress level 1 (\emph{weak stress}) or
2 (\emph{strong stress}),
and $s_c$ is the stress category, which can be \emph{study, work, family, interpersonal relation, romantic relation,} or \emph{self-cognition}.
We aggregated user $u$'s stressful periods into the total number of stress periods,
the average stress level, and the number of different stress categories
that user $u$ experienced throughout the year.

\vspace{1.2mm}
$Stress(u)$=$(sNum(u), sLevel(u), sCatNum(u)),$

\vspace{1.2mm}
\noindent
where $sNum(u)$ = $|{\cal S}(u)|$,
$sLevel(u)$ = $\frac{\sum _{s \in {\cal S}(u)} s_l}{|{\cal S}(u)|}$, and

$~~~~sCatNum(u)$ = $|\{s_c~|~s \in {\cal S}(u)\}|$.

Furthermore, people who are suffering from psychological disorders like depression or
bipolar disorder are also more prone to suffering from suicidal ideation.
People who previously tried to attempt suicide also tend to
have a higher suicidal risk than those who did not.
Like \cite{coppersmith2014quantifying}, we set
$Disorder(u)$=1, if and only if user $u$ posted something on microblog like
\emph{``I have depression/bipolar disorder $\cdots$"}, and 0 otherwise.
$Attempt(u)$=1, if and only if user $u$ posted something on microblog like
\emph{``I've committed suicide \#number\# times $\cdots$"}, and 0 otherwise.

\subsubsection{Personality (Pursuing Perfection, Ruminant Thinking, Interpersonal Sensitive)}

Involvement of personality factors in susceptibility to suicidality
has been the subject of research since the 1950s.
Psychological findings showed that personality dimensions are significantly associated
with suicide-related behaviors, and
specific personality factors impact suicide uniquely for each gender~\cite{bluml2013personality}.
Such personality types like 
\emph{pursuing perfection, ruminant thinking,} and \emph{interpersonal sensitivity}
could be useful markers of suicide risk~\cite{greenspon2014there}.

To assess the extent of one's \emph{pursuing perfection} characteristic, we
identified 36 potential perfectionist users from the
``Perfectionism" subreddit (\url{https://www.reddit.com/r/perfectionism/}),
whose posts contained sentences like ``\emph{like many perfectionists, I feel ...}", ``\emph{I definitely have unhealthy perfectionism}", and so on. We calculated top 100 high-frequency words from their posts. We
remove stop words and words representing things like examination.
and acquire such words like ``\emph{no enough}", ``\emph{perfect}",
 ``\emph{loser}", ``\emph{failure}", ``\emph{imperfect}", ``\emph{compete}", and so on.
We took them as seed words, and enrich the perfection-related lexicon with synonymous or similar words.
We also made a ruminant thinking-related lexicon consisting of
very destructive negative words such as
``\emph{regret}", ``\emph{repent}", ``\emph{rue}", ``\emph{penitent}",
``\emph{self-blame}", etc.
Table~\ref{tab:perfectRuminantLexicon} lists typical perfection-related and ruminant thinking-related words that we used in the study.

\begin{table}[htbp]
\centering
\caption{\centering Typical perfection-related and ruminant thinking-related words.}
\begin{tabular}{p{1.2cm} p{6.6cm}}
\hline\noalign{\smallskip}
\textbf{Type}  &    \textbf{Words}   \\
\noalign{\smallskip}\hline\noalign{\smallskip}
Perfection-Related &  perfect, perfectionist, perfection, stupid, wrong, upset, mood, struggle, goal, realistic, not enough, lose, loser, failure, imperfect, compete, force, negative, disappoint, despair, letdown, useless, always, never, still, upset, problem
  \\
\hline
Ruminant Thinking-Related & regret, repent, rue, penitent, confess, hate, self-blame,
grievance, complaint, injustice, unfairness, growl, discontent, lose, loser, owe, sinner, die \\
\hline\noalign{\smallskip}
\end{tabular}%
\label{tab:perfectRuminantLexicon}%
\end{table}

\begin{figure*}[htp]
    \centering
    \includegraphics[scale=0.86]{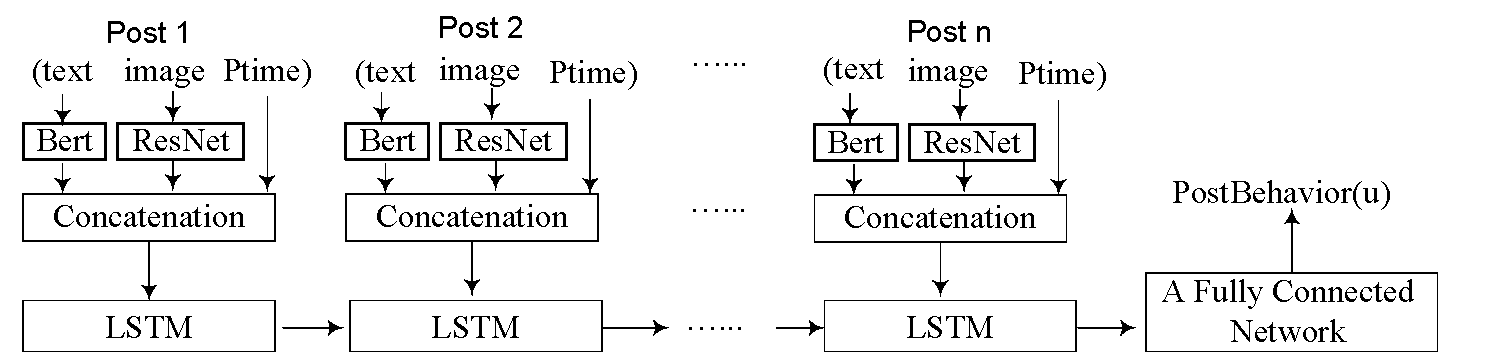}
    \caption{The framework for learning user $u$'s post behavior.}
    \label{fig:texts-LSTM}
\end{figure*}

We measured user's pursuing perfection and ruminant thinking based on
the mean proportion of perfection-related and ruminant-related words per post, respectively.
Let $P(u)$ denote the set of posts made by user $u$.
For each post $p \in P(u)$, assume
function $pwordNum(p,u)$ and $rwordNum(p,u)$ return the number of perfection-related
and ruminant-related words in $p$, respectively.

\vspace{1.2mm}
$Perfect(u)$ = $\frac{1}{|P(u)|}$ $\sum_{p \in P(u)} \frac{pwordNum(u,p)}{|p|}$

\vspace{1.2mm}
$Ruminat(u)$ = $\frac{1}{|P(u)|}$ $\sum_{p \in P(u)} \frac{rwordNum(u,p)}{|p|}$
\vspace{1.2mm}

We measured user $u$'s interpersonal sensitivity by counting how many times
the user experienced stressful periods in the stress category of \emph{inter-personnel relation}.

\vspace{1.2mm}
$Sensitive(u)$ =

$~~~~~~|\{s~|~s \in {\cal S}(u) \wedge s_c = ``interpersonal~relation"\}|$

\vspace{1.2mm}
\noindent where ${\cal S}(u)$ is a set of stressful periods detected
from user $u$'s posts, and  $s$=$(s_p, s_l, s_c)$ is one of it in ${\cal S}(u)$.

\subsubsection{Post Behavior (Text, Image, Post Time)}

An individual may post texts and images anytime at will.
These linguistic and/or visual contents, as well as the posting time,
may reveal personal thoughts or feelings.
Since Bert \cite{xiao2018bertservice} and ResNet \cite{he2016deep} are common used pre-trained models which perform well on textual feature extraction \cite{du2020adversarial,chen2020distilling} and visual feature extraction \cite{zha2020adversarial,dai2019deep}.
For each post, we encoded its linguistic and visual contents
into a 768-dimensional and a 300-dimensional vector, respectively,
through a pre-trained Bert model~
and a 34-layer ResNet.
When the image was missing, we filled in with
a default null image, which is usually displayed at the position for
image wanted in applications.
When the post contains multiple images, we took the average feature vector
of the images as the visual content representation of the post.
We mapped the user's absolute post time into the hour from 0 to 23.
Assume user $u$ made totally $n$ posts
$post_1 (text_1, image_1, ptime_1), \cdots,$ $post_n (text_n, image_n,$ $ptime_n)$, where
$text_i \in \mathbb{R}^{768 \times 1}$, $image_i \in \mathbb{R}^{300 \times 1}$, and
$ptime_i \in \{0,1,2,\cdots, 23\}$ for $1 \leq i \leq n$.

To further learn user $u$'s posts representation, we
concatenated ($text_i, image_i$, $ptime_i$) as a
1069-dimensional vector $postvec_i$, where
$postvec_i$=$text_i||image_i||ptime_i \in  \mathbb{R}^{1 \times 1069}$, and then
applied LSTM to extract textual and visual information from
$postvec_1,\cdots, postvec_n$.
Let the hidden dimension size of LSTM be 300.
\[out_i, h_i = LSTM(postvec_i, h_{i-1}).\]
The last output of LSTM was $out_n \in \mathbb{R}^{1 \times 300}$.
To reduce the computational complexity of subsequent calculations, a fully connected layer was
used to generate a 30-dimensional vector as the value of user's \emph{post behavior} property:
\[PostBehavior(u) = ReLU(out_n ~W_0+b_0) \in \mathbb{R}^{1 \times 30},\]

\noindent where $out_n \in \mathbb{R}^{1 \times 300}, W_0 \in \mathbb{R}^{300 \times 30}$, and $b_0 \in \mathbb{R}^{1 \times 30}$.

Details of learning user $u$'s post behavior representation are illustrated in Figure~\ref{fig:texts-LSTM}.

\begin{table*}
\caption{Statistics of users' personal suicide-oriented knowledge graphs on Weibo.}
\label{tab:sta-kg}
\centering
\begin{tabular}{|p{1.5cm}<{\centering}|p{6.5cm}| p{4cm}| p{4cm}|}
\hline 
\textbf{Category} & \textbf{Properties} & \textbf{Users with Suicidal Ideation} & \textbf{Ordinary Users without Suicidal Ideation} \\
\hline
            & Gender (\emph{male,female,unknown}) & (21.3\%, 78.5\%, 0.2\%) & (50.6\%, 42.3\%, 7.1\%) \\ \cline{2-4}
Personal & Age on average & 25.8 & 28.3 \\ \cline{2-4}
Information & Location (\emph{east, south, north, south-west, north-west, middle, north-east, unknown})
            & (18.8\%, 10.5\%, 7.4\%, 6.6\%, 3.0\%, 7.0\%, 3.3\%, 43.5\%)
            & (25.0\%, 15.0\%, 15.7\%, 5.8\%, 2.9\%, 7.9\%, 5.3\%, 22.3\%) \\ \hline
            & Pursuing Perfection & 0.25\%  & 0.17\% \\ \cline{2-4}
Personality & Ruminant Thinking   & 0.086\% & 0.052\%\\ \cline{2-4}
            & Interpersonal Sensitive & 1.3 & 1.0 \\ \hline
            & Number of Stressful Periods & 2.1 & 1.8 \\ \cline{2-4}
Experience  & Psychological Disorder & 0.1\% & 0.03\% \\ \cline{2-4}
            & Previous Attempt &3.5\% & 0.1\% \\ \hline
Post        & Number of Texts & 252,901 & 491,130 \\ \cline{2-4}
Behavior    & Number of Images & 93,461  & 260,667\\ \hline
            & Proportion of Suicide Words Per Post& 0.034\% & 0.016\% \\ \cline{2-4}
Emotion     & Proportion of Last Words Per Post & 0.0013\% & 0.000023\% \\ \cline{2-4}
Expression  & Proportion of Future Words Per Post & 0.031\% & 0.045\% \\ \cline{2-4}
            & Proportion of Negation Words Per Post & 0.012\% & 0.009\% \\ \cline{2-4}
            & Proportion of Self-Concern Words Per Post & 0.079\% & 0.029\%\\ \cline{2-4}
            & Emotion Transition ($love$-$joy$, $love$-$anxiety/sorrow$)  & (0.1, 0.5) &  (0.3, 0.1) \\  \hline
            & Following Number  & 207.0 & 378.1 \\ \cline{2-4}
Social      & Follower Number  & 566.9 & 1515.3\\ \cline{2-4}
Interaction & Mention/Forward/Agree Number  & 3.4 & 10.9 \\ \cline{2-4}
            & Number of Neighbour Users  & 4.3  & 5.1 \\
\hline\noalign{\smallskip}
\end{tabular}

\end{table*}

\subsubsection{Emotion Expression (Suicide Words, Last Words, Future Words, Negation Words, Self-Concerns, Emotion Transition)}

Suicides tend to express rather than repress their despair suicidal feelings,
less future words, more negation and
self-referenced first-person pronouns such as \emph{I, me, my, myself,} and \emph{mine}, etc.
in their posts~\cite{ML15}.
The existence of last words (which express regret, guilty,
wishes towards their families and friends, or funeral),
particularly in those recent posts, 
is also recognized as an important observable risk factor of suicidal ideation~\cite{JS69}.
From the Chinese Suicide Dictionary~\cite{lv2015creating} and Chinese Linguistic Inquiry and Word Count \cite{huang2012development}, we extracted 586 about suicide, 125 about last words,
86 about future words, 665 about negation words, and 36 words/phrases about self-concern.
We calculated the mean word frequencies (i.e., mean word proportions over the total words per post)
as the property values of $SuicideProp(u)$, $LastWordProp(u)$, $FutureProp(u)$, $NegProp(u)$,
and $SelfProp(u)$ for user $u$.
Here, only two most recent weeks' posts were considered in the computation of $LastWordProp(u)$.


Apart from the specific words, we drew the inspiration from the study~\cite{FR16}
that there is a significant difference in emotion transition between
suicide users and non-suicide users. That is,
love emotion is more regularly followed by anxiety or sorrow emotion among the suicide people,
while love is more regularly followed by joy emotion among the non-suicide people.
Hereby, as in~\cite{FR16}, we considered eight emotions (\emph{love, joy, expectation, anxiety, sorrow, anger, hate,} and \emph{surprise}), and counted two types of emotion transitions
from love to joy ($love$-$joy(u)$) or from love to anxiety/sorrow ($love$-$anxiety/sorrow(u)$) within the
ten most recent continuous posts as the property value of emotion transition:
$EmotionTrans(u)$ = $(love$-$joy(u), love$-$anxiety/sorrow(u))$.

For user $u$, if there existed a pair of posts, where the first one contained
one or more love lexicons, and the second later post contained one or more joy lexicons
based on the Chinese emotion lexicons DUTIR (http://ir.dlut.edu.cn/),
the value of $love$-$joy(u)$ was increased by 1.
The value of  $love$-$anxiety/sorrow(u)$ was computed in the same way.


\subsubsection{Interaction (Following Number, Follower Number, Mention/Forward/Agree Number, Neighbour Users)}

Social isolation is a significant and reliable predictor of suicidal ideation.
When trapped in trouble, if one could get the support,
understanding, and acceptance from family, friends and peers,
his/her stress and follow-up negative emotions could be offset, which could probably avoid extreme suicidal behavior.
In reality, desperate people tend to have a weak social network, and thus get weak social supports~\cite{LM14}.
Here, one's online social engagement is reflected from his
following/follower number, mention/forward/agree number,
and neighbour users.
Specially, in order not to be affected by ``Zombie fan" (fake fans in social media), we filtered out fans who posted less than three original posts in the last six months when counting the follower number.

$Interact(u)$ = $(FollowingNum(u)$, $FollowerNum(u)$, $IntActNumb(u)$, $Users(u))$,
where the first three elements take non-negative integer values,
and $Users(u)$
is a set of neighbour users which user $u$ follows.

Formally, we describe the personal suicide-oriented knowledge graph
as $G=\{V, E\}$,
where $V$ and $E$ are node set and edge set respectively.
A user node can have multiple property nodes, 
and may link to another user node via the following relationship
on the social media.
The weight of a social interaction-user edge implies the suicidal emotion influence between the two users
on the social media,
and the weight of a user-property edge represents the contribution of
the property to the suicidal ideation.
These weights will be computed during the following learning process.

Finally, the statistic of knowledge graph is shown in Table \ref{tab:sta-kg}.

%% file: new-Sec3-Methods-GNN.tex
\subsection{Suicidal Ideation Detection based on the Personal Knowledge Graph}
We designed a property attention mechanism and neighbour attention mechanism
to reason the key factors related to suicidal ideation,
and discover the suicidal emotion influence among the neighbour users
on the social media.

\subsubsection{Property Attention}

Excepting user $u$'s neighbour users,
we concatenated all the instantiated properties from $u$'s personal knowledge graph,
and obtained a 61-dimensional property vector
$P_u$ = $(p_1,p_2,\cdots,p_n) \in \mathbb{R}^{61 \times 1}$,
where $n$=61.

To compute the significance of different properties,
we enforced $P_u$ with a property attention vector $\alpha \in \mathbb{R}^{1 \times 61}$
and acquired a new property vector $P'_u \in \mathbb{R}^{61 \times 1}$:
\begin{equation}\nonumber
\begin{aligned}
P'_u &= P_u \times \alpha^T, \\
\alpha &= softmax((P_u)^{T}W_{1}+b_1),
\end{aligned}
\end{equation}
\noindent where $W_{1} \in \mathbb{R}^{61 \times 61}$ and $b_{1} \in \mathbb{R}^{1 \times 61}$ are
two trainable parameters.
\begin{figure}[htp]
    \centering
    \includegraphics[scale=0.6]{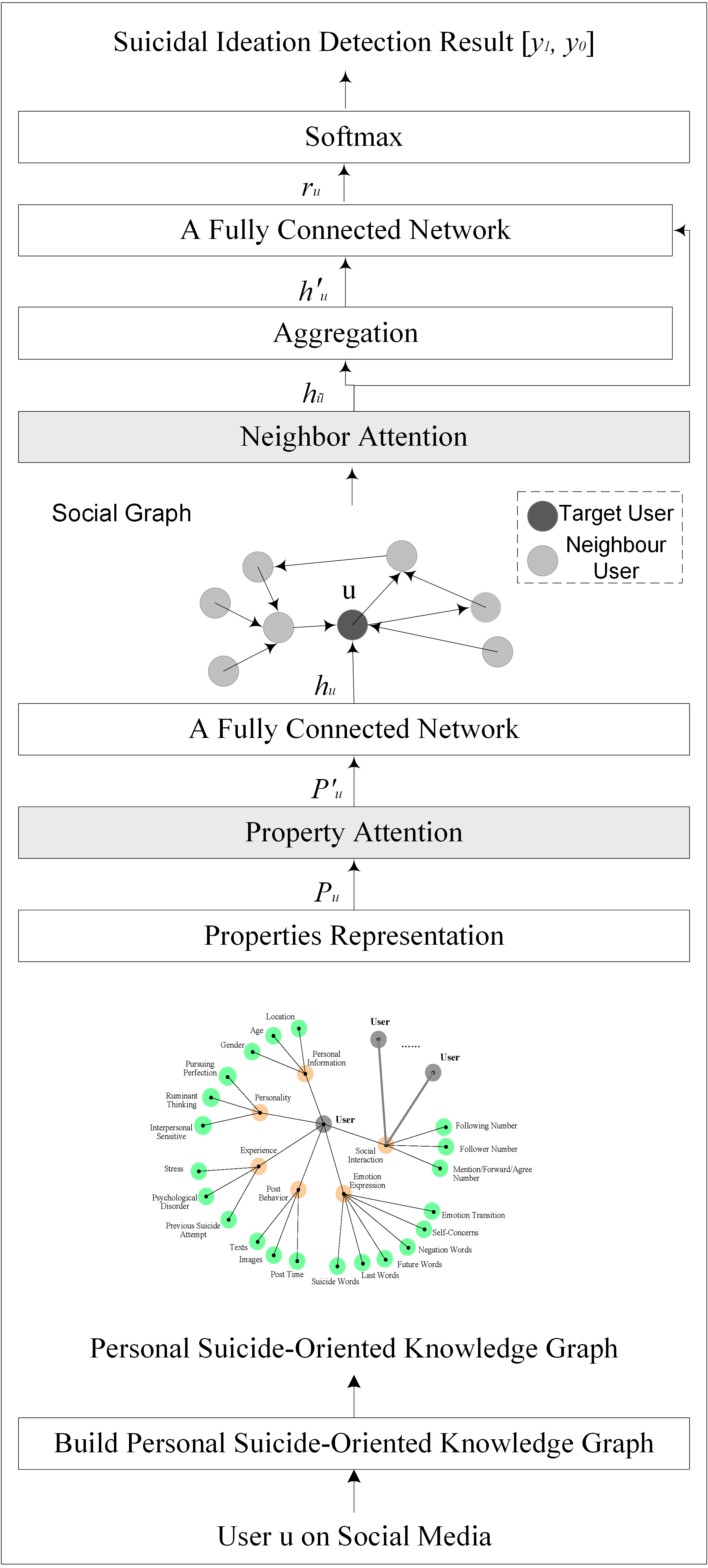}
    \caption{ Architecture of the personal knowledge graph based suicidal ideation detection.}
    \label{fig:model}
\end{figure}


\subsubsection{Neighbour Attention}

We imposed neighbour influences upon user $u$'s property representation $P'_u$.
Considering different neighbour influences upon user $u$'s suicidal ideation,
for instance, some people may easily be influenced by elders, and else may trust their peers more,
we adopted a neighbour attention mechanism similar to the
graph attention mechanism~\cite{DBLP:conf/iclr/VelickovicCCRLB18}.

To reduce computational complexity, we firstly figured out user
$u$'s initial hidden state $h_u \in \mathbb{R}^{1 \times 60} $ from his property representation
$P'_u$ through a fully connected layer:
\[h_u = tanh((P'_u)^T ~W_{2}+b_{2}),\]
\noindent
where $W_{2} \in \mathbb{R}^{61 \times 60}$ and $b_{2} \in \mathbb{R}^{1 \times 60}$ are trainable parameters.

Let $\mathbb{N}_u$ be the set of direct neighbour users linked with user $u$
on the social media.
For each neighbour pair $(u,\widetilde{u})$ (where $\widetilde{u} \in \mathbb{N}_u$),
we obtained the attention coefficient $c_{u,\widetilde{u}} \in \mathbb{R}^{1 \times 1}$
through another fully connected layer with concatenate operation $||$:
\[c_{u,\widetilde{u}} = tanh((h_u||h_{\widetilde{u}}) ~W_{3}+b_{3}),\]
\noindent where $W_{3} \in \mathbb{R}^{120 \times 1}$ and $b_{3} \in \mathbb{R}^{1 \times 1}$ are two trainable parameters.

In this way, we obtained a vector of user $u$'s neighbour attention coefficients:
\[C_u= [c_{u,1},c_{u,2},\cdots,c_{u,|\mathbb{N}_u|}] \in \mathbb{R}^{1 \times |\mathbb{N}_u|}\]
\noindent
where $|\mathbb{N}_u|$ is the number of $u$'s direct neighbours.

A softmax function was then applied to compute a series of scores
${\cal B}_u = [\beta_{u,1},\beta_{u,2},\cdots,\beta_{u,|\mathbb{N}_u|}] \in \mathbb{R}^{1 \times |\mathbb{N}_u|}$ to represent the influences of neighbour users $\mathbb{N}$ upon user $u$.
The higher the score is, the bigger the suicidal influence is.
\[{\cal B}_u = softmax(C_u).\]

By means of ${\cal B}_u$, we aggregated information from neighbour users and
updated $u$'s hidden state $h_u$ accordingly into
$h'_u \in \mathbb{R}^{1 \times 60}$:
\[h'_u = \sigma(\sum_{\widetilde{u} \in \mathbb{N}_u}~\beta_{u,\widetilde{u}} ~h_{\widetilde{u}} + h_u).\]

Once obtained, we utilized a fully connected layer to get the user's final representation
$r_u \in \mathbb{R}^{1 \times 60}$:
\[r_u = tanh(h'_u~W_{4} + b_{4}),\]
\noindent
where $W_{4} \in \mathbb{R}^{60 \times 60}$ and $b_{4} \in \mathbb{R}^{1 \times 60}$.

\subsubsection{Suicidal Ideation Detection}

Finally, a fully connected layer and softmax function were applied
to detect suicidal ideation of user $u$:
\[[y_1,y_0] = softmax(r_u~W_{5} +b_{5}),\]
\noindent
where $y_{1},y_{0}$ represent the possibility of user $u$ with or without suicidal ideation, $W_{5} \in \mathbb{R}^{60 \times 2}$ and $b_{5} \in \mathbb{R}^{1 \times 2}$ are trainable parameters.

%% file: new-Sec4-Results-new.tex
\section{Experiments}

\subsection{Datasets}

In this study, we evaluate our personal suicide-oriented knowledge graph based suicidal ideation detection methods
on two types of datasets: microblog posts from Sina Weibo and forum posts from Reddit.

\subsubsection{Sina Weibo Dataset}




On March 17, 2012, a user which screen name was ``Zoufan"
left his last word on Sina Weibo and then committed suicide.
Since then over 160,000 users gathered here and posted more than
1,700,000 comments, and the numbers keep growing today,
forming a large microblog tree hole.
The majority of these commenting posts spoke out the posters' minds,
disclosing their tragic experiences, hopeless thoughts, and even plans of suicide.

We crawled all the users' commenting posts from May 1, 2018 to April 30, 2019,
and computed each user's expression degrees related to suicide according to
the Chinese social media-based suicide dictionary~\cite{ML15}.
The dictionary lists 2168 suicide-related words and phrases, falling into 13
categories (suicidal thought, self-harm, psychache, mental illness, hopeless feeling,
somatic complain, self-regulation, negative personality, stress,
trauma/hurt, talking about others, shame/guilt, anger/hostility)
Each word/phrase is assigned a weight from 1 to 3,
indicating its binding degree with suicide risk.


Assume user $u$ posted a set of comments $P(u)$ in the tree hole.
We define $u$'s expression degree related to suicide by counting his total
suicide-related weighted words/phases in $P(u)$.

\vspace{1.2mm}
$UserDegree(u)$ = $~\sum_{post \in P(u)} postDegree(post)$
\vspace{1.2mm}

$postDegree(post)~~$ = $~\sum_{w \in post} weight(w)$
\vspace{1.2mm}

\noindent
where $weight(w)$ is the weight of word/phrase $w$ in $p$
according to the suicide dictionary~\cite{ML15}. A word/phrase not in the suicide dictionary has weight 0.

We ranked and selected top 4,000 users who had high expression degrees related to
suicide. After that, we recruited four PhD researchers to manually scan the 4,000 users and keep
3,652 users, who clearly mentioned suicide thought, suicide plan, or self-injury
at least five times in different days.
For example, if a user expressed clear suicidal thoughts like \emph{``At this moment, I especially want to die. I feel very tired. I really want to be free."}, \emph{``Heh, even Weibo cant give me a reason to keep on going."} or \emph{``I don't want to do anything for the last five days of my life."} more than 5 times in different days, then we label him/her at suicide risk.
Finally, we labeled these 3,652 users with suicidal ideation at suicide risk.

As comparison, we randomly selected 3,677 ordinary users on Sina Weibo, subject to the following three conditions.
Firstly, they were active users, each making over 100 normal posts on the open microblog
during the same temporal period.
Secondly, neither of their posts contained any suicide-related word or phrase.
Thirdly, they did not make any commenting post in any hidden tree hole.
In the study, we regarded the ordinary users without suicidal ideation.
If two users had a following-follower relationship on Sina Weibo,
we built an edge linking the two neighbour users.
Table~\ref{tab:MicroblogData} illustrates the data collected from Sina Weibo.

\begin{table*}
\caption{Statistic of data collected from Sina Weibo.}
\label{tab:MicroblogData}
\centering
\begin{tabular}{p{4.2cm} m{1.6cm}<{\centering} m{2.1cm}<{\centering} m{3.6cm}<{\centering} m{4.2cm}<{\centering}}
\hline\noalign{\smallskip}
    & \textbf{\# Users}  & \textbf{\# Normal Posts}  & \textbf{\# Normal Posts with Images} & \textbf{\# Neighbour Users to be followed}  \\
\noalign{\smallskip}\hline\noalign{\smallskip}
  Users with suicidal ideation            & 3,652  & 252,901  & 93,461  & 4.3 \\
  Ordinary users without non-suicide      & 3,677  & 491,130  & 260,667 & 5.1  \\
  Total users                             & 7,329  & 744,031  & 354,128 & 4.7 \\
  \hline\noalign{\smallskip}
\end{tabular}
\end{table*}

The training set, validation set, and test set contained
6,129, 600, and 600 users, who were exclusively and randomly chosen from the total 7,329 microblog users,
respectively.
Users with suicidal ideation and ordinary users without suicidal ideation
occupied nearly half in each set.

\subsubsection{Reddit Dataset}
The Reddit dataset was crawled from the popular online forum Reddit, and contained 500 users in total.
There were five categories of users, i.e.,
Supportive, Suicide Indicator, Suicidal Ideation, Suicidal Behavior, and Actual Attempt
with user number of
108, 99, 171, 77, and 45, respectively.
From Supportive to Actual Attempt, the risk of suicide gradually increases.
We split the Reddit dataset into training set, validation set, and test set with the number of 303, 99, and 98, respectively.

\subsection{Experimental Setup}
We compared the performance of our personal suicide-oriented knowledge graph based method
with the following four methods.
\begin{itemize}

    \item \textbf{CNN}~\cite{gaur2019knowledge}: A Convolutional Neural Network takes embeddings of user posts as input.
    \item \textbf{LSTM+Attention}~\cite{coppersmith2018natural}: An self-attention mechanism based Long Short-Term Memory model which captures the contextual information between suicide-related words and others.
    \item \textbf{SDM}~\cite{lei2019}: An hierarchical attention network based on Long Short-Term Memory module and ResNet module.
    \item \textbf{Text+History+Graph}~\cite{Pradyumna2019a}: A multipronged approach includes bidirectional Long Short-Term Memory module and Graph Neural Network.
    The textual content shared in Twitter, the historical tweeting activity of the users and social network formed between different users posting about suicidality are concerned to identify and explore users' suicidal ideation.
\end{itemize}

As Reddit dataset only contained posts from each user, during the experiment,
we deleted the ResNet module and the user profile module of the SDM method,
the Social Graph module of the Text+History+Graph method,
as well as the Neighbour Attention module, Personal Information property and Social Interaction property of our KG-based method. We took Bert contextual word embeddings~\cite{xiao2018bertservice} as our word embeddings.

Five metrics (accuracy, precision, recall, F1-measure, macro-F1-measure) were adopted in the performance evaluation.

\vspace{1.2mm}
$Precision$ = $\frac{TP}{TP+FP},~~~~$
$Recall$ = $\frac{TP}{TP+FN}$

\vspace{1.2mm}
$Accuracy$ = $\frac{TP+TN}{TP+FP+TN+FN}$

\vspace{1.2mm}
$F1$-$measure$ = $2 \times \frac{Precision \times Recall}{Precision+Recall}$

\vspace{1.2mm}
\noindent where true positive (TP), false positive (FP), true negative
(TN), and false negative (FN) are defined in the following confusion matrix (Table~\ref{tab:ConfusionMatrix}).

As toward Reddit dataset there were more than two classes (actually 5),
we employed a commonly used multi-class macro-F1-measure, which
calculates F1-measure for each class, and then finds their un-weighted mean.

\begin{table}[!htbp]
\caption{\centering  Confusion matrix used to evaluate the performance.}
\label{tab:ConfusionMatrix}
\centering
\begin{tabular}{|l|l|l|}
\hline
\diagbox{Actual}{Detected}
                   &  Positive  & Negative   \\ \hline
Positive   &  True Positive (TP)  & False Negative (FN)  \\ \hline
Negative    &  False Positive (FP)  & True Negative (TN)   \\ \hline
\end{tabular}
\end{table}

\subsection{Effectiveness of Involving Personal Suicide-Oriented Knowledge Graphs in Suicidal Ideation Detection}

Table~\ref{tab:KG-Performance} shows that, with the considerations of visual information from posts and users' profile information, the SDM method achieved better performance compared with that of CNN method and LSTM-Attention method.
After using the social graph of users, Text+History+Graph method outperformed the above methods slightly in accuracy and F1-measure. Furthermore, involving personal suicide-oriented knowledge graphs into
suicidal ideation detection could improve the detection accuracy, F1-measure, precision, and recall by
2.18\%, 1.88\% , 1.9\%, 1.87\% over the Text+History+Graph method,
demonstrating the effectiveness of the personal knowledge graph based method.

Since five-class classification is a harder task than two-class classification,
and the modalities of the Reddit dataset are less than that of the Sina Weibo dataset,
the overall performance of all the methods dropped, compared with that on the Sina Weibo dataset, as shown
in Table~\ref{tab:KG-Performance-reddit}.
Without the contribution of social graph, the performance of the
Text+History+Graph method dropped to 59.61\% and 59.02\% in accuracy and F1-measure, respectively,
which were worse than the SDM method.
With the help of personal suicide-oriented knowledge graph,
the KG-based method still outperformed the rest methods.

\begin{table}
\caption{\centering Performance Comparison on the Sina Weibo Dataset.}
\label{tab:KG-Performance}
\centering
\begin{tabular}{p{2.6cm} p{1cm} p{1cm} p{1cm} p{1cm}}
\hline\noalign{\smallskip}
\textbf{Method}     & \textbf{Acc.}     & \textbf{F1.}  & \textbf{Prec.}  & \textbf{Rec.} \\
\noalign{\smallskip}\hline\noalign{\smallskip}
CNN & 86.77\% & 84.47\% & 84.19\% & 84.77\% \\
LSTM+Attention & 88.89\% & 88.10\% & 88.56\% & 87.65\% \\
SDM & 91.35\% & 90.97\% & 90.11\%& 91.85\% \\
Text+History+Graph & 91.56\% & 91.81\% & 91.85\% & 91.77\% \\
KG-based method & \textbf{93.74\%} &  \textbf{93.69\%}  &\textbf{93.75\%}  & \textbf{93.64\%}\\
\hline\noalign{\smallskip}
\end{tabular}
\end{table}

\begin{table}
\caption{\centering Performance Comparison on the Reddit dataset.}
\label{tab:KG-Performance-reddit}
\centering
\begin{tabular}{p{2.6cm} p{1cm} p{1cm} p{1cm} p{1cm}}
\hline\noalign{\smallskip}
\textbf{Method}     & \textbf{Acc.}     & \textbf{macro.F1.}  & \textbf{Prec.}  & \textbf{Rec.} \\
\noalign{\smallskip}\hline\noalign{\smallskip}
CNN                & 52.31\%  & 52.61\% & 53.05\% & 52.19\% \\
LSTM+Attention     & 55.12\% & 54.49\% & 53.16\% & 55.89\% \\
SDM                & 60.58\%  & 60.44\% & 60.31\% & 60.58\% \\
Text+History+Graph & 59.61\%  & 59.02\% & 58.42\% & 59.64\% \\
KG-based method & \textbf{65.92\%} &  \textbf{65.93\%}  &\textbf{65.35\%}  & \textbf{66.21\%}\\
\hline\noalign{\smallskip}
\end{tabular}
\end{table}

\begin{table}[t]
\caption{\centering Performance of property and neighbour attention mechanisms.}
\label{tab:ablation_att}
\centering
\begin{tabular}{p{3.5cm} p{0.8cm} p{0.8cm} p{0.8cm} p{0.8cm}}
\hline\noalign{\smallskip}
\textbf{Method}     & \textbf{Acc.}     & \textbf{F1.}  & \textbf{Prec.}  & \textbf{Rec.} \\
\noalign{\smallskip}\hline\noalign{\smallskip}
avg-weight-property      & 92.55\%   & 93.23\%    &93.55\%  & 92.91\%     \\
avg-weight-neighbour     & 92.17\%   & 93.02\%    & 93.17\%  &92.88\%     \\
random-weight-property   & 90.12\%   & 89.56\%    & 89.49\%  & 89.63\%      \\
random-weight-neighbour  & 91.86\%   & 92.16\%    & 92.55\%  & 91.77\%      \\
property+neighbour attention  & \textbf{93.74\%}  & \textbf{93.69\%} & \textbf{93.75\%}  & \textbf{93.64\%}   \\
\hline\noalign{\smallskip}
\end{tabular}
 \end{table}

\begin{table}[ht]
\caption{ Performance of using different graph neural network (GNN) models.}
\label{tab:GNNs}
\centering
\begin{tabular}{p{3.5cm} p{0.8cm} p{0.8cm} p{0.8cm} p{0.8cm}}
\hline\noalign{\smallskip}
\textbf{Method}     & \textbf{Acc.}     & \textbf{F1.}  & \textbf{Prec.}  & \textbf{Rec.} \\
\noalign{\smallskip}\hline\noalign{\smallskip}
GCN                & 92.26\%   & 92.69\% & 92.86\%  & 92.54\%      \\
GraphSAE         & 92.49\%   & 91.16\% & 90.89\%  & 91.43\%      \\
GAT-inspired  & \textbf{93.74\%}  & \textbf{93.69\%} & \textbf{93.75\%}  & \textbf{93.64\%}   \\
(property+neighbour attention) & & & & \\
\hline\noalign{\smallskip}
\end{tabular}
 \end{table}

\subsection{Effectiveness of Property and Neighbour Attention Mechanisms}

Property attention and neighbour attention are two crucial modules of our personal knowledge graph based
method. They aim to reason about the key properties and find the most influential neighbour users to user's suicidal ideation, reflected from their associated property weights and neighbour weights, respectively.
To investigate the effectiveness of the two attention mechanisms,
we replaced each learnt attention weight with an average or random value in the range of [0,1], respectively.
Table~\ref{tab:ablation_att} shows that the two attention mechanisms were effective, achieving the best performance
compared to the alternative average and random weight solutions, and
the average solution was better than the random solution.

It is worth mentioning here that the property+neighbour attention strategy drew
inspirations from the Graph Attention networks (GAT)~\cite{DBLP:conf/iclr/VelickovicCCRLB18}, which incorporates
the neighbor attention mechanism. This also contributes to its best performance than those of the other two
well-known GNN models (GCN \cite{kipf2016semi} and GraphSAGE \cite{hamilton2017inductive}), which are commonly used
for neighbour information aggregation.
As illustrated in Table~\ref{tab:GNNs},
after replacing the neighbour attention module with the GCN method or GraphSAGE method,
there are obvious declines about more than 1.25\%, in  accuracy  and  1\% in
F1-measure on the Weibo dataset,
since the neighbour attention mechanism can derive the different influence from one's different neighbour users.


\subsection{Contributions of Social Neighbours and Personal Suicide-Oriented Knowledge Graph to Suicidal Ideation Detection}



The effectiveness of the personal suicide-oriented knowledge graph prompted us to further
investigate the impact of different personal factors
upon suicidal ideation detection.
We used \emph{information gain} to assess the change of information amount involving
a certain personal factor (property)
based on \emph{entropy} and \emph{conditional entropy}.
A bigger information gain implies a bigger significance of the factor.

Formally, let $F$ be a category, and $f$ be a vector value in the domain of category $F$, i.e.,
$f \in Dom(F)$.
Let $y \in \{y_1,y_0\}$ denote the detection result (with/without suicidal ideation).
$Prob(.)$ represents probability and $H(.)$ is the entropy.

\vspace{1.2mm}
$InfoGain(y|F)$ =  $H(y)- H(y|F)$

\vspace{1.2mm}
\noindent where
$H(y)$=$-\sum_{y \in \{y_1,y_0\}} Prob(y) ~log Prob(y),$

\vspace{1.2mm}
$~~~~H(y|F)$=$\sum_{f \in Dom(F)} Prob(F$=$f) ~H(y|F$=$f),$

\vspace{1.2mm}
$~~~~H(y|f)$=$\sum_{y \in \{y_1,y_0\}} Prob(y|F$=$f)~log Prob(y|F$=$f)$.
\vspace{1.2mm}

We firstly considered the six categories of personal properties (\emph{Personal Information,
Personality, Experience, Post Behaviors, Emotion Expression}, and
\emph{Social Interaction}) in one's personal suicide-oriented knowledge graph,
and then delved into the concrete properties of the key categories.

To ease the computation,
for a property which takes continuous values (like  \emph{Age}, \emph{Pursuing Perfection}, \emph{Ruminant Thinking}, \emph{Interpersonal Sensitive}, \emph{Stress Duration}, \emph{Stress Level}, \emph{Stress Category},
\emph{Suicide Words}, \emph{Last Words}, \emph{Future Words}, \emph{Negation Words},
\emph{Self-Concerns Words},
\emph{Following Number}, \emph{Follower Number}, and \emph{Mention/Forward/Agree Number}),
we used the mean to divide all its values into two classes,
and transformed the domain of the property into $\{0,1\}$.
The domains of discrete property values (like \emph{Location}, \emph{Gender} and \emph{Emotion Transition})
remained unchanged.

\begin{figure}
    \centering
   \includegraphics[scale=0.33]{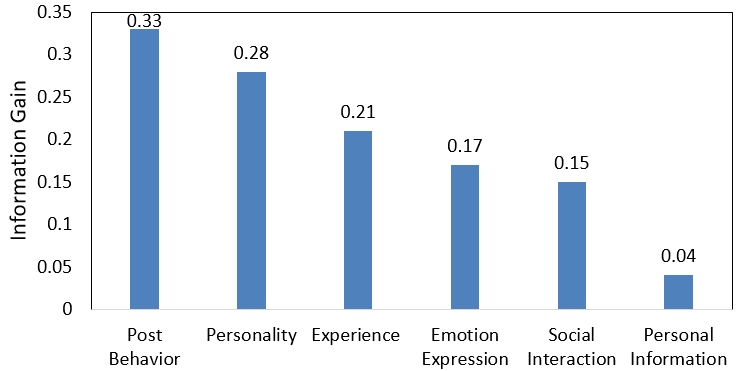}
    \caption{Information gains of the top-3 categories.}
    \label{fig:CategoryContribution}
\end{figure}

\begin{figure*}
    \centering
    \includegraphics[scale=0.42]{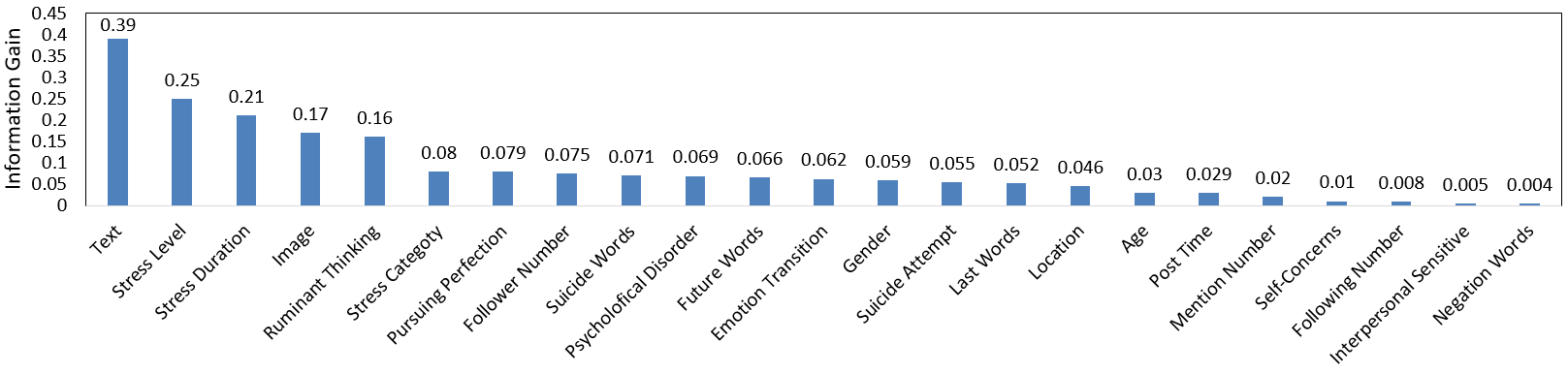}
    \caption{ Information gains of all the properties from personal suicide-oriented knowledge graph.}
    \label{fig:FactorContribution}
\end{figure*}

\begin{figure}
    \centering
    \includegraphics[scale=0.4]{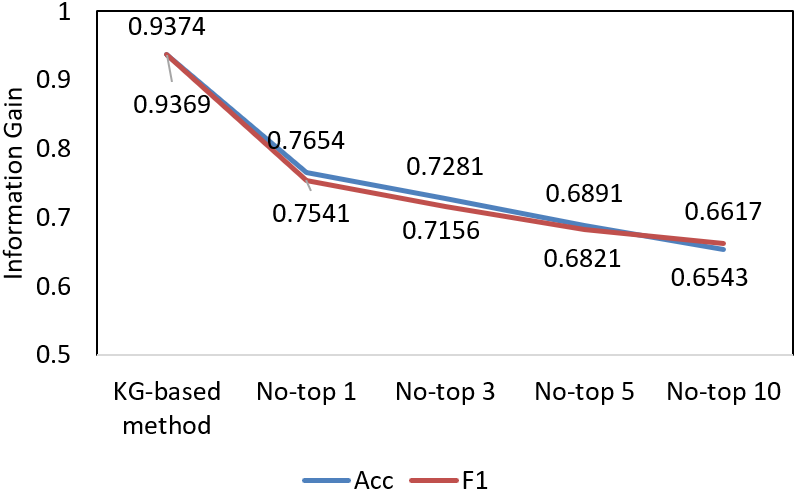}
    \caption{Results of the feature selection experiment through
    removing the top-$x$ key properties in information gain (where $x$=1,3,5,10).}
    \label{fig:feature-sele}
\end{figure}

We mapped the \emph{Texts} property values under the \emph{Post Behavior} category
to three classes based on the emotional polarities given by SnowNLP (https://github.com/isnowfy/snownlp).
Assume user $u$ wrote a sequence of texts $Texts(u)$ in his posts.

{
\[Class(Texts(u)) = \\
\left\{\begin{array}{ll}
0 & \mbox{if $EP(Texts(u)) \leq -0.3$;} \\
1 & \mbox{if $-0.3 < EP(Texts(u)) < 0.3$;} \\
2 & \mbox{if $EP(Texts(u)) \geq 0.3$}
\end{array}
\right.\]
}

\vspace{1.2mm}
\noindent where $EP(Texts(u))=\frac{\sum_{t \in Texts(u)}  SnowNLP(t)}{|Texts(u)|}$.

In a similar way, we mapped the \emph{Images} property values under the \emph{Post Behavior} category
to four classes based on brightness and proportion of warm color, calculated by RGB.
Assume user $u$ posted a sequence of images $Images(u)$.
{
\[Class(Images(u)) = \\
\left\{\begin{array}{ll}
0 & \mbox{if $B(Images(u)) < 0.5$ $\wedge$} \\
  & \mbox{$~~~~W(Images(u)) < 0.5$;} \\
1 & \mbox{if $B(Images(u)) < 0.5$ $\wedge$} \\
  & \mbox{$~~~~W(Images(u)) \geq 0.5$;} \\
2 & \mbox{if $B(Images(u)) \geq 0.5$ $\wedge$} \\
  & \mbox{$~~~~W(Images(u) < 0.5$;} \\
3 & \mbox{if $B(Images(u)) \geq 0.5$ $\wedge$} \\
  & \mbox{$~~~~W(Images(u)) \geq 0.5$}
\end{array}
\right.\]
}

\vspace{1.2mm}
\noindent where
$B(Images(u))=\frac{\sum_{i \in Images(u)}  RGB-Bright(i)}{|Images(u)|}$, and
$~~~~~~~~~~~~~~W(Images(u))=\frac{\sum_{i \in Images(u)}  RGB-Warm(i)}{|Images(u)|}$.

Figure~\ref{fig:CategoryContribution} shows the average information gain of each
category in the detection of suicidal ideation. From
the result presented, we can see that all the listed categories
brought positive impact on suicidal detection,
and the top-3 categories were \emph{Post Behavior},
\emph{Personality}, and \emph{Experience}, whose information gains were
0.33, 0.28, and 0.21, respectively.

We further looked into all the categories, and examined the contributions from
their respective properties. As shown in Figure~\ref{fig:FactorContribution},
the top-5 key properties were \emph{posted text, stress level, stress duration,} \emph{image,} and
\emph{ruminant thinking}.


\begin{table*}[]
\caption{Derived Five Test datasets Containing Different Proportions of Users With Anti-Real Posts.}
\label{tab:testDataSet}
\centering
\begin{tabular}{p{5cm} p{2cm} p{2cm} p{2cm} p{2cm} p{2cm}}
\noalign{\smallskip}\hline\noalign{\smallskip}
 & $TD_1$ & $TD_2$ & $TD_3$ & $TD_4$ & $TD_5$  \\ \noalign{\smallskip}\hline\noalign{\smallskip}
\# Users with suicidal ideation & 43 & 100 & 200 & 300 & 257  \\ 
~~\# with anti-real posts         & 43 (100\%) & 43 (43\%) & 43 (21.5\%) & 43 (14.3\%) & 0 (0\%)   \\
~~\# without anti-real posts      & 0  & 57  & 157 & 257 & 257  \\  \hline
\# Ordinary users (random selection) & 43 & 100 & 200 & 300 & 257 \\  
\hline\noalign{\smallskip}
\end{tabular}
\end{table*}

To be more illustrative in the detection of suicidal ideation, we conducted a further feature selection experiment.
From the result presented in Figure~\ref{fig:feature-sele}, we note that
after removing the top-$x$ key properties in information gain (where $x$=1,3,5,10),
both accuracy and F1-measure kept declining, verifying the effectiveness of the key properties
in suicidal risk analysis.

\subsection{Impact of Data Noise on Suicidal Ideation Detection Performance}

\begin{table}[]
\caption{Detection Performance on the Five Derived Test Datasets
Containing Different Proportions of Users with Anti-Real Posts.}
\label{tab:anti_real}
\centering
\begin{tabular}{p{2.4cm} p{0.8cm} p{0.8cm} p{0.8cm} p{0.8cm} p{0.8cm}}
\hline\noalign{\smallskip}
\textbf{Method} & \textbf{$TD_1$} &\textbf{$TD_2$} & $TD_3$ & $TD_4$ & $TD_5$  \\ 
\noalign{\smallskip}\hline\noalign{\smallskip}
CNN                & 65.75\%     & 73.73\%     & 82.97\%     & 86.77\% & 90.12\%  \\
LSTM+Attention     & 67.27\%     & 76.54\%     & 84.23\%     & 88.89\% & 92.12\%  \\
SDM                & 74.98\%     & 82.31\%     & 87.59\%     & 91.35\% & 94.03\%  \\
Text+History+Graph & 75.61\%     & 83.59\%     & 88.64\%     & 91.56\% & 94.06\%  \\ \noalign{\smallskip}\hline\noalign{\smallskip}
KG-based method    & \textbf{80.08\%}     & \textbf{88.11\%}     & \textbf{90.76\%}    & \textbf{93.74\%} & \textbf{95.45\%}  \\
--On suicidal users & 66.51\%      & 82.33\%       & 88.06\%       & 93.97\%      & 97.58\%       \\
--On ordinary users & 93.65\%      & 93.89\%       & 93.46\%       & 93.51\%      & 93.51\%   \\
\hline\noalign{\smallskip}
\end{tabular}
\end{table}

Due to the less restrictive and free-style nature of the social media,
the presence of data noise is quite common, and may have a misleading effect on the experimental results.
To examine the impact of data noise on the suicidal ideation detection performance,
we consider two types of data noise: (1) \textbf{anti-real posts} (that is, users didn't fill in real information of themselves on the social media.) (2) \textbf{a few posts} (that is, users made only a few posts on the social media),
and conducted experiments on a few datasets, each possessing different levels of data noise.

\subsubsection{Users with Anti-Real Posts}

Thanks to the existence of the tree hole, we could glimpse some anti-real sentiment expression
from a user's normal open post as follows.
For a user who made posts in both the tree hole and on the open microblog platform, 
if (1) he expressed despair, depression, anxiety, or suicidal ideation in the tree hole,
but showed uplift feelings on the open posts within the 24 hours;
and (2) the above situation happened more than twice within the latest one month, 
we annotated the user with anti-real posts.

Our test Sina Weibo dataset contains 300 users with suicidal ideation and 300 ordinary users.
From the 300 users with suicidal ideation, we identified totally 43 users with anti-real posts.
By varying the proportions of users with suicidal ideation in the test dataset,
we derived 5 test datasets with different levels of data noise.  
In each derived test dataset (Table~\ref{tab:testDataSet}), 
as ordinary users made no posts in the tree hole,
we randomly picked up the same number of ordinary users 
from the 300 ordinary test users as the control group.

Table~\ref{tab:anti_real} lists the performance of all the methods on the five different test datasets.
Our knowledge graph based method outperforms all the other methods in the four metrics. 
The less proportions of the users with anti-real posts, the higher the detection performance.
When the proportion of the users with suicidal ideation and anti-real posts dropped to 21.5\%, 14.3\%, and 0\%,
the detection accuracy of the knowledge graph based method on the suicidal group reached 88.06\%,
93.97\%, and 97.58\%, respectively. 
The detection performance for the ordinary group remained unchanged, around 93.65\% in accuracy and
93.89\% in F1-measure.
 
\subsubsection{Users with Only a Few Posts}

From the total 600 test users (300 ordinary users and 300 users with suicidal ideation),
we filtered out 88 users with suicidal ideation and another 88 ordinary
users without suicidal ideation, and all of them had less than 5 posts throughout the year (Table~\ref{tab:testDataSet6}).

\begin{table}[]
\caption{Derived Test Dataset Containing Users With Less Than 5 Posts throughout the Year.}
\label{tab:testDataSet6}
\centering
\begin{tabular}{p{6cm} p{2cm} }
\hline\noalign{\smallskip}
 & \textbf{$TD_6$}   \\ \noalign{\smallskip}\hline\noalign{\smallskip}
\# Users with suicidal ideation & 88  \\
\# Ordinary users & 88   \\
\hline\noalign{\smallskip}
\end{tabular}
\end{table}

\begin{table}
\caption{Detection Performance on the Derived Test Dataset
Containing Users with Only a Few Posts.}
\label{tab:fewPosts}
\centering
\begin{tabular}{p{2.6cm} p{1cm} p{1cm} p{1cm} p{1cm}}
\hline\noalign{\smallskip}
\textbf{Method}     & \textbf{Acc.}     & \textbf{F1.}  & \textbf{Prec.}  & \textbf{Rec.} \\
\noalign{\smallskip}\hline\noalign{\smallskip}
CNN                & 72.35\%  & 72.38\% & 72.12\% & 72.65\% \\
LSTM+Attention     & 74.56\% & 74.91\% & 74.98\% & 74.84\% \\
SDM                & 85.49\%  & 85.36\% & 85.46\% & 85.26\% \\
Text+History+Graph & 86.88\%   &86.52\%  &86.49\%   & 86.55\% \\ \noalign{\smallskip}\hline\noalign{\smallskip}
KG-based method & \textbf{89.73\%} &  \textbf{89.49\%}   & \textbf{89.92\%} & \textbf{89.49\%} \\
--On suicidal users &    92.04\%   &   90.00\%    &     88.04\%   &    92.04\%   \\
--On ordinary users &    87.50\%   &  89.53\%      &  91.67\%      &    87.50\%  \\
\hline\noalign{\smallskip}
\end{tabular}
\end{table}

It is not surprising that the performance of all the trained models (in Table~\ref{tab:fewPosts})
dropped, given such a small number of test posts. Among the five detection methods, 
our knowledge graph based method achieved the highest overall accuracy of 89.73\% on all the users,   
92.04\% on the suicidal group, and 87.50\% on the ordinary group.

\begin{table}[t]
\caption{Performance on the New Dataset.}
\label{tab:new_dataset}
\centering
\begin{tabular}{p{2.6cm} p{1cm} p{1cm} p{1cm}  p{1cm}}
\hline\noalign{\smallskip}
\textbf{Method}        & \textbf{Acc.}  & \textbf{F1.}   & \textbf{Prec.}  & \textbf{Rec.}  \\
\noalign{\smallskip}\hline\noalign{\smallskip}
CNN                 & 87.56\%  & 87.65\% & 87.61\% & 87.69\%  \\
LSTM+Attention      & 88.86\%  & 87.99\% & 88.23\% & 87.77\% \\
SDM                 & 91.78\%  & 90.87\% & 90.28\% & 91.47\% \\
Text+History+Graph  & 92.01\%   &91.62\% & 91.76\% & 91.48\%     \\
KG-based method     & \textbf{94.53\%} &  \textbf{94.31\%} & \textbf{94.45\%} & \textbf{94.18\%} \\
\hline\noalign{\smallskip}
\end{tabular}
\end{table}

\subsubsection{Experiments on a Small-Sized New Dataset}

Apart from examining the impact of data noise through the six datasets derived from our Weibo test dataset,
we also built a small-sized new dataset, containing 85 suicide users (who had passed away as verified 
by news from 2012 to 2014) and another 85 ordinary users randomly sampled from our test dataset.
It is expected that the data noise of this new dataset was less than that of our Sina Weibo dataset.
 Comparatively, suicide users in this new dataset had more supportive information.

(1) They made more posts (average 83.2 in their last year) than the suicidal ideation users in our
Weibo dataset (average 69.2 in the latest one year).

(2) They experienced more stressful periods (average 6.0), 
showed more negative expressions (23\% posts with negative emotions given by SnowNLP) 
in the last year than the suicidal ideation users (average 69.2 posts, 2.1 stressful periods, 12\% posts with negative emotion in the latest one year).

With the help of the supportive information,  
our knowledge graph based method could deliver a very high performance,
94.53\% in accuracy and 94.33\% in F1-measure, as shown in Table~\ref{tab:new_dataset}.

To sum up, the proposed knowledge graph based suicidal ideation detection model
outperforms the other baseline methods, when facing challenging few-posts and anti-real-posts on microblog.

\subsection{An Ablation Study}

As the introduction of the user knowledge graph is essential in this study,
we conducted an ablation study to see the performance with only the user's post behavior (Without-KG method).
As shown in Table~\ref{tab:KG-Performance-ablation}, \ref{tab:KG-Performance-reddit-ablation}
there are huge declines about 4.15\%, 7.58\% in accuracy and 4.48\%, 6.91\% in F1-measure on both datasets.
The mediocre performance of the Without-KG method introduced the effectiveness of personal suicide-oriented knowledge graph.

\begin{table}[t]
\caption{\centering Results of the Ablation Study on the Sina Weibo Dataset.}
\label{tab:KG-Performance-ablation}
\centering
\begin{tabular}{p{2.6cm} p{1cm} p{1cm} p{1cm} p{1cm}}
\hline\noalign{\smallskip}
\textbf{Method}     & \textbf{Acc.}     & \textbf{F1.}  & \textbf{Prec.}  & \textbf{Rec.} \\
\noalign{\smallskip}\hline\noalign{\smallskip}

KG-based method & \textbf{93.74\%} &  \textbf{93.69\%}  &\textbf{93.75\%}  & \textbf{93.64\%}\\
Without-KG method & 89.59\%  & 89.21\% & 89.68\% & 88.74\% \\
\hline\noalign{\smallskip}
\end{tabular}
\end{table}
\begin{table}[ht]
\caption{\centering Results of the ablation study on the Reddit Dataset.}
\label{tab:KG-Performance-reddit-ablation}
\centering
\begin{tabular}{p{2.6cm} p{1cm} p{1cm} p{1cm} p{1cm}}
\hline\noalign{\smallskip}
\textbf{Method}     & \textbf{Acc.}     & \textbf{macro.F1.}  & \textbf{Prec.}  & \textbf{Rec.} \\
\noalign{\smallskip}\hline\noalign{\smallskip}
KG-based method & \textbf{65.92\%} &  \textbf{65.93\%}  &\textbf{65.35\%}  & \textbf{66.21\%}\\
Without-KG method & 58.34\%  & 59.02\% & 59.41\% & 58.63\% \\
\hline\noalign{\smallskip}
\end{tabular}
\end{table}

\subsection{Training and Inference Time Costs}

Table~\ref{tab:time} shows the training time costs of the five methods until convergence
(getting the best result on the validation set) on a computer server with 4 NVIDIA GTX 1080Ti GPU.
As our KG-based method utilized the most modalities (including text, image, social graph, and knowledge graph),
its training time is the most.
CNN requires the least training time, as it does not use the time-consuming sequential module.
The inference time cost per batch (batch size is 16) of all the five methods are close except for CNN.

\begin{table}[]
\caption{Training and Inference Time Costs of the Methods on Sina Weibo Dataset.}
\label{tab:time}
\centering
\begin{tabular}{p{2.7cm} m{1.8cm}<{\centering} m{2.7cm}<{\centering} }
\hline\noalign{\smallskip}
\textbf{Method}     & \textbf{Training Time}     & \textbf{Inference Time(batch)}  \\
\noalign{\smallskip}\hline\noalign{\smallskip}
CNN (text)               & 56.7 mins    & 37.9 seconds  \\ 
LSTM+Attention (text)     & 282.4 mins   & 49.6 seconds \\ 
SDM (text+image)               & 362.3 mins   & 55.4 seconds \\ 
Text+History+Graph (text+graph) & 286.8 mins   & 50.8 seconds \\ 
KG-based method (text+image+graph+KG)    & 386.5 mins   & 56.4 seconds\\
\hline\noalign{\smallskip}
\end{tabular}
\end{table}

%% file: new-Sec5-Discussion.tex
\section{Discussions}
\subsection{Ethical Considerations}
Keeping ethical considerations in mind is essential for the task of suicidal ideation detection.
All the data (including name, age, posts, etc.) was crawled from the public social media,
and was only used for research. We anonymized the data before labeling.
There was no interaction or intervention with the subjects.

\subsection{Limitations}

\subsubsection{Insufficiency of Microblog Data}
While the detection result is promising,
the construction of users' personal suicide-oriented knowledge graphs, particularly personality-related factors, is still very preliminary in the current study, leaving a few important factors out of the study
due to resource limitations.
For instance, according to the psychological study~\cite{ParentingStyle1993},
individual's \emph{parenting rearing style}
is a persistent factor that affects an individual's mental health.
It is widely believed that
individuals who are poorly educated are more likely to develop suicidal ideation
than individuals who are actively educated. Parental rearing style is of great significance
to the formation and development of individual's personality. Positive parental rearing style like
emotional warmth and understanding, rather than negative excessive interference and
excessive protection, contributes to one's healthy growth~\cite{ParentingStyle2001}.
Completely relying on social media to analyze such personal knowledge
is limited, and some other channels need to be explored to identify
whether one was positively or negatively brought up.

\subsubsection{Noise of Microblog data}
 
Due to the less restrictive and free-style nature of social media like microblog, it is quite possible that someone may have a few more accounts, or may not want to fill in real information on the social media, such as Weibo. 
In this study, we empirically examined the impact of data noise on suicidal ideation detection performance.
Two types of data noise were considered: (1) users didn't fill in real information of themselves on the social media;
and (2) users made only a few posts on the social media.
We conducted experiments on a few derived datasets and another new dataset, and the results 
were expected. That is, the lower data noise exhibits, the higher detection performance can be achieved. 
This raised a very interesting question: ``\emph{can we firstly take a look at the data first before
feeding them to train a detection model?}" 
In addition, 
user identification and inference of users real thoughts and feelings will also be quite desirable, deserving further investigation.

%% file: new-Sec6-Conclusion.tex
\section{Conclusion}

In this paper, we built and used a personal suicide-oriented knowledge graph
for suicidal ideation detection on social media.
A two-layered attention mechanism was deployed to explicitly reason and establish key risk factors to individuals' suicidal ideation.
The performance study on
7,329 microblog users (3,652 with suicidal ideation and 3,677 without suicidal ideation)
show that: 1) with the constructed personal knowledge graph, the social media-based
suicidal ideation detection can achieve over 93\% accuracy, outperforming the state-or-art approach;
and 2) among the six categories of personal factors, \emph{post, personality,} and \emph{experience} were the
top-3 key indicators. Under these categories, \emph{posted text}, \emph{stress level},
\emph{stress duration}, \emph{posted image}, and \emph{ruminant thinking} contribute the
most to one's suicidal ideation detection.

While the paper shows some promising results of the proposed method,
leveraging social media to accurately identify personal properties is still limited.
More reliable data sources and domain-specific expert knowledge
need to be used and integrated into suicidal ideation detection.
Dynamic maintenance of user-centric knowledge graph 
also deserves deep investigation for building really handy solutions.

%% file: Main.bbl
\begin{thebibliography}{10}
\providecommand{\url}[1]{#1}
\csname url@samestyle\endcsname
\providecommand{\newblock}{\relax}
\providecommand{\bibinfo}[2]{#2}
\providecommand{\BIBentrySTDinterwordspacing}{\spaceskip=0pt\relax}
\providecommand{\BIBentryALTinterwordstretchfactor}{4}
\providecommand{\BIBentryALTinterwordspacing}{\spaceskip=\fontdimen2\font plus
\BIBentryALTinterwordstretchfactor\fontdimen3\font minus
  \fontdimen4\font\relax}
\providecommand{\BIBforeignlanguage}[2]{{%
\expandafter\ifx\csname l@#1\endcsname\relax
\typeout{** WARNING: IEEEtran.bst: No hyphenation pattern has been}%
\typeout{** loaded for the language `#1'. Using the pattern for}%
\typeout{** the default language instead.}%
\else
\language=\csname l@#1\endcsname
\fi
#2}}
\providecommand{\BIBdecl}{\relax}
\BIBdecl

\bibitem{Nock2008}
M.~Nock, G.~Borges, E.~Bromet, and et~al., ``Cross-national prevalence and risk
  factors for suicidal ideation, plans and attempts,'' \emph{British Journal of
  Psychiatry}, vol. 192, no.~2, pp. 98--105, 2008.

\bibitem{Jacob2014}
N.~Jacob, J.~Scourfield, and R.~Evans, ``Suicide prevention via the
  {Internet},'' \emph{J. of Crisis}, 2014.

\bibitem{bagge1998suicide}
C.~Bagge and A.~Osman, ``The suicide probability scale: Norms and factor
  structure,'' \emph{Psychological reports}, vol.~83, no.~2, pp. 637--638,
  1998.

\bibitem{fu2007predictive}
K.-w. Fu, K.~Y. Liu, and P.~S. Yip, ``Predictive validity of the chinese
  version of the adult suicidal ideation questionnaire: Psychometric properties
  and its short version.'' \emph{Psychological Assessment}, vol.~19, no.~4, p.
  422, 2007.

\bibitem{harris2015abc}
K.~M. Harris, J.-J. Syu, O.~D. Lello, Y.~E. Chew, C.~H. Willcox, and R.~H. Ho,
  ``The abc’s of suicide risk assessment: Applying a tripartite approach to
  individual evaluations,'' \emph{PLoS One}, vol.~10, no.~6, p. e0127442, 2015.

\bibitem{E-health2014}
H.~Christensen, P.~Batterham, and B.~O'Dea, ``E-health interventions for
  suicide prevention,'' \emph{International journal of environmental research
  and public health}, vol.~11, no.~8, pp. 8193--8212, 2014.

\bibitem{essau2005frequency}
C.~A. Essau, ``Frequency and patterns of mental health services utilization
  among adolescents with anxiety and depressive disorders,'' \emph{Depression
  and anxiety}, vol.~22, no.~3, pp. 130--137, 2005.

\bibitem{rickwood2007and}
D.~J. Rickwood, F.~P. Deane, and C.~J. Wilson, ``When and how do young people
  seek professional help for mental health problems?'' \emph{Medical journal of
  Australia}, vol. 187, no.~S7, pp. S35--S39, 2007.

\bibitem{zachrisson2006utilization}
H.~D. Zachrisson, K.~R{\"o}dje, and A.~Mykletun, ``Utilization of health
  services in relation to mental health problems in adolescents: a population
  based survey,'' \emph{BMC public health}, vol.~6, no.~1, p.~34, 2006.

\bibitem{ji2020suicidalb}
S.~Ji, S.~Pan, X.~Li, E.~Cambria, G.~Long, and Z.~Huang, ``Suicidal ideation
  detection: A review of machine learning methods and applications,''
  \emph{IEEE Transactions on Computational Social Systems}, 2020.

\bibitem{Alambo2019}
A.~Alambo, M.~Gaur, U.~Lokala, U.~Kursuncu, K.~Thirunarayan, A.~Gyrard,
  A.~Sheth, R.~S. Welton, and J.~Pathak, ``Question answering for suicide risk
  assessment using reddit,'' in \emph{2019 IEEE 13th International Conference
  on Semantic Computing (ICSC)}.\hskip 1em plus 0.5em minus 0.4em\relax IEEE,
  2019, pp. 468--473.

\bibitem{Cheng2017Assessing}
Q.~Cheng, T.~M. Li, C.~L. Kwok, T.~Zhu, and P.~S. Yip, ``Assessing suicide risk
  and emotional distress in chinese social media: a text mining and machine
  learning study,'' \emph{Journal of Medical Internet Research}, vol.~19,
  no.~7, p. e243, 2017.

\bibitem{Du2018}
J.~Du, Y.~Zhang, J.~Luo, Y.~Jia, Q.~Wei, C.~Tao, and H.~Xu, ``Extracting
  psychiatric stressors for suicide from social media using deep learning,''
  \emph{BMC medical informatics and decision making}, vol.~18, no.~2, p.~43,
  2018.

\bibitem{SawhneyMMSS18}
R.~Sawhney, P.~Manchanda, P.~Mathur, R.~Shah, and R.~Singh, ``Exploring and
  learning suicidal ideation connotations on social media with deep learning,''
  in \emph{Proc. 9th Workshop on Computational Approaches to Subjectivity,
  Sentiment and Social Media Analysis}, 2018, pp. 167--175.

\bibitem{coppersmith2018natural}
G.~Coppersmith, R.~Leary, P.~Crutchley, and A.~Fine, ``Natural language
  processing of social media as screening for suicide risk,'' \emph{Biomedical
  informatics insights}, vol.~10, p. 1178222618792860, 2018.

\bibitem{Vioul2018Detection}
M.~J. Vioul\'{e}s, B.~Moulahi, A.~J., and S.~Bringay, ``Detection of
  suicide-related posts in twitter data streams,'' \emph{{IBM} Journal of
  Research \& Development}, vol.~62, no.~1, pp. 7:1--7:12, 2018.

\bibitem{Robinson2015}
J.~Robinson, M.~Rodrigues, S.~Fisher, E.~Bailey, and H.~Herrman, ``Social media
  and suicide prevention: findings from a stakeholder survey,'' \emph{Shanghai
  Arch. Psychiatry}, vol.~27, no.~9, pp. 27--35, 2015.

\bibitem{Madelyn2003}
M.~Gould, P.~Jamieson, and D.~Romer, ``Media contagion and suicide among the
  young,'' \emph{American Behavioral Scientist}, vol.~46, no.~9, pp.
  1269--1284, 2003.

\bibitem{gaur2019knowledge}
M.~Gaur, A.~Alambo, J.~P. Sain, U.~Kursuncu, K.~Thirunarayan, R.~Kavuluru,
  A.~Sheth, R.~Welton, and J.~Pathak, ``Knowledge-aware assessment of severity
  of suicide risk for early intervention,'' in \emph{The World Wide Web
  Conference}.\hskip 1em plus 0.5em minus 0.4em\relax ACM, 2019, pp. 514--525.

\bibitem{Brisset2014}
L.~Brisset, Y.~Leanza, E.~Rosenberg, B.~Vissandj\'{e}e, L.~J. Kirmayer,
  G.~Muckle, S.~Xenocostas, and H.~Laforce, ``Language barriers in mental
  health care: A survey of primary care practitioners,'' \emph{Journal of
  immigrant and minority health}, vol.~16, no.~6, pp. 1238--1246, 2014.

\bibitem{Nie2015}
L.~Nie, Y.-L. Zhao, M.~Akbari, J.~Shen, and T.-S. Chua, ``Bridging the
  vocabulary gap between health seekers and healthcare knowledge,'' \emph{IEEE
  Transactions on Knowledge and Data Engineering}, vol.~27, no.~2, pp.
  396--409, 2015.

\bibitem{ning2017knowledge}
G.~Ning, Z.~Zhang, and Z.~He, ``Knowledge-guided deep fractal neural networks
  for human pose estimation,'' \emph{IEEE Transactions on Multimedia}, vol.~20,
  no.~5, pp. 1246--1259, 2017.

\bibitem{xue2019knowledge}
F.~Xue, R.~Hong, X.~He, J.~Wang, S.~Qian, and C.~Xu, ``Knowledge based topic
  model for multi-modal social event analysis,'' \emph{IEEE Transactions on
  Multimedia}, 2019.

\bibitem{chaudhary2019enhancing}
C.~Chaudhary, P.~Goyal, D.~N. Prasad, and Y.-P.~P. Chen, ``Enhancing the
  quality of image tagging using a visio-textual knowledge base,'' \emph{IEEE
  Transactions on Multimedia}, vol.~22, no.~4, pp. 897--911, 2019.

\bibitem{limsopatham-collier-2016-normalising}
L.~Nut and C.~Nigel, ``Normalising medical concepts in social media texts by
  learning semantic representation,'' in \emph{Proceedings of the 54th Annual
  Meeting of the Association for Computational Linguistics}, 2016, pp.
  1014--1023.

\bibitem{Pradyumna2019a}
P.~P. Sinha, R.~Mishra, R.~Sawhney, D.~Mahata, R.~R. Shah, and H.~Liu,
  ``\#suicidal - a multipronged approach to identify and explore suicidal
  ideation in {Twitter},'' in \emph{Proc. 28th ACM International Conference on
  Information and Knowledge Management}, 2019.

\bibitem{Profile2019}
A.~Mbarek, S.~Jamoussi, A.~Charfi, and A.~B. Hamadou, ``Suicidal profiles
  detection in {Twitter},'' in \emph{Proc. of the 15th Intl. Conf. on Web
  Information Systems and Technologies (WEBIST)}, 2019, pp. 289--296.

\bibitem{DBLP:conf/iclr/VelickovicCCRLB18}
\BIBentryALTinterwordspacing
P.~Velickovic, G.~Cucurull, A.~Casanova, A.~Romero, P.~Li{\`{o}}, and
  Y.~Bengio, ``Graph attention networks,'' in \emph{6th International
  Conference on Learning Representations, {ICLR} 2018, Vancouver, BC, Canada,
  April 30 - May 3, 2018, Conference Track Proceedings}.\hskip 1em plus 0.5em
  minus 0.4em\relax OpenReview.net, 2018. [Online]. Available:
  \url{https://openreview.net/forum?id=rJXMpikCZ}
\BIBentrySTDinterwordspacing

\bibitem{pestian2017machine}
J.~P. Pestian, M.~Sorter, B.~Connolly, K.~Bretonnel~Cohen, C.~McCullumsmith,
  J.~T. Gee, L.-P. Morency, S.~Scherer, L.~Rohlfs, and S.~R. Group, ``A machine
  learning approach to identifying the thought markers of suicidal subjects: a
  prospective multicenter trial,'' \emph{Suicide and Life-Threatening
  Behavior}, vol.~47, no.~1, pp. 112--121, 2017.

\bibitem{crawford2003depression}
J.~R. Crawford and J.~D. Henry, ``The depression anxiety stress scales (dass):
  Normative data and latent structure in a large non-clinical sample,''
  \emph{British journal of clinical psychology}, vol.~42, no.~2, pp. 111--131,
  2003.

\bibitem{henry2005short}
J.~D. Henry and J.~R. Crawford, ``The short-form version of the depression
  anxiety stress scales (dass-21): Construct validity and normative data in a
  large non-clinical sample,'' \emph{British journal of clinical psychology},
  vol.~44, no.~2, pp. 227--239, 2005.

\bibitem{just2017machine}
M.~A. Just, L.~Pan, V.~L. Cherkassky, D.~L. McMakin, C.~Cha, M.~K. Nock, and
  D.~Brent, ``Machine learning of neural representations of suicide and emotion
  concepts identifies suicidal youth,'' \emph{Nature human behaviour}, vol.~1,
  no.~12, p. 911, 2017.

\bibitem{harris2017suicide}
K.~M. Harris and M.~T.-T. Goh, ``Is suicide assessment harmful to participants?
  findings from a randomized controlled trial,'' \emph{International journal of
  mental health nursing}, vol.~26, no.~2, pp. 181--190, 2017.

\bibitem{braithwaite2016validating}
S.~R. Braithwaite, C.~Giraud-Carrier, J.~West, M.~D. Barnes, and C.~L. Hanson,
  ``Validating machine learning algorithms for twitter data against established
  measures of suicidality,'' \emph{JMIR mental health}, vol.~3, no.~2, p. e21,
  2016.

\bibitem{harris2013}
K.~M. Harris, J.~P. McLean, and J.~Sheffield, ``Suicidal and online: How do
  online behaviors inform us of this high-risk population?'' \emph{Death
  studies}, vol.~38, no.~6, pp. 387--394, 2014.

\bibitem{JP10}
J.~Pestian, H.~Nasrallah, P.~Matykiewicz, A.~Bennett, and A.~Leenaars,
  ``Suicide note classification using natural language processing: A content
  analysis,'' \emph{Biomedical informatics insights}, vol.~3, pp. BII--S4706,
  2010.

\bibitem{YH07}
Y.-P. Huang, T.~Goh, and C.~L. Liew, ``Hunting suicide notes in web
  2.0-preliminary findings,'' in \emph{Ninth IEEE International Symposium on
  Multimedia Workshops (ISMW 2007)}.\hskip 1em plus 0.5em minus 0.4em\relax
  IEEE, 2007, pp. 517--521.

\bibitem{MT13}
T.~M. Li, B.~C. Ng, M.~Chau, P.~W. Wong, and P.~S. Yip, ``Collective
  intelligence for suicide surveillance in web forums,'' in \emph{Pacific-Asia
  Workshop on Intelligence and Security Informatics}.\hskip 1em plus 0.5em
  minus 0.4em\relax Springer, 2013, pp. 29--37.

\bibitem{NM14}
N.~Masuda, I.~Kurahashi, and H.~Onari, ``Suicide ideation of individuals in
  online social networks,'' \emph{PloS one}, vol.~8, no.~4, p. e62262, 2013.

\bibitem{MC16}
M.~De~Choudhury, E.~Kiciman, M.~Dredze, G.~Coppersmith, and M.~Kumar,
  ``Discovering shifts to suicidal ideation from mental health content in
  social media,'' in \emph{Proceedings of the 2016 CHI conference on human
  factors in computing systems}.\hskip 1em plus 0.5em minus 0.4em\relax ACM,
  2016, pp. 2098--2110.

\bibitem{shing2020prioritization}
H.-C. Shing, P.~Resnik, and D.~W. Oard, ``A prioritization model for
  suicidality risk assessment,'' in \emph{Proceedings of the 58th Annual
  Meeting of the Association for Computational Linguistics}, 2020, pp.
  8124--8137.

\bibitem{jones2019analysis}
N.~Jones, N.~Jaques, P.~Pataranutaporn, A.~Ghandeharioun, and R.~Picard,
  ``Analysis of online suicide risk with document embeddings and latent
  dirichlet allocation,'' in \emph{2019 8th International Conference on
  Affective Computing and Intelligent Interaction Workshops and Demos
  (ACIIW)}.\hskip 1em plus 0.5em minus 0.4em\relax IEEE, 2019, pp. 1--5.

\bibitem{tadesse2020detection}
M.~M. Tadesse, H.~Lin, B.~Xu, and L.~Yang, ``Detection of suicide ideation in
  social media forums using deep learning,'' \emph{Algorithms}, vol.~13, no.~1,
  p.~7, 2020.

\bibitem{JJ14}
J.~Jashinsky, S.~H. Burton, C.~L. Hanson, J.~West, C.~Giraud-Carrier, M.~D.
  Barnes, and T.~Argyle, ``Tracking suicide risk factors through twitter in the
  {US},'' \emph{Crisis}, 2014.

\bibitem{HS15}
H.~Sueki, ``The association of suicide-related twitter use with suicidal
  behaviour: a cross-sectional study of young internet users in {Japan},''
  \emph{Journal of affective disorders}, vol. 170, pp. 155--160, 2015.

\bibitem{FR16}
F.~Ren, X.~Kang, and C.~Quan, ``Examining accumulated emotional traits in
  suicide blogs with an emotion topic model,'' \emph{J. of biomedical and
  health informatics}, vol.~20, no.~5, pp. 1384--1396, 2015.

\bibitem{LG14}
L.~Guan, B.~Hao, and T.~Zhu, ``How did the suicide act and speak differently
  online? behavioral and linguistic features of china's suicide microblog
  users,'' \emph{arXiv preprint arXiv:1407.0466}, 2014.

\bibitem{LZ15}
L.~Zhang, X.~Huang, T.~Liu, A.~Li, Z.~Chen, and T.~Zhu, ``Using linguistic
  features to estimate suicide probability of chinese microblog users,'' in
  \emph{Proceedings of International Conference on Human Centered Computing},
  2015, pp. 549--559.

\bibitem{huang2015topic}
X.~Huang, X.~Li, T.~Liu, D.~Chiu, T.~Zhu, and L.~Zhang, ``Topic model for
  identifying suicidal ideation in chinese microblog,'' in \emph{Proceedings of
  the 29th Pacific Asia Conference on Language, Information and Computation},
  2015, pp. 553--562.

\bibitem{guan2015identifying}
L.~Guan, B.~Hao, Q.~Cheng, P.~S. Yip, and T.~Zhu, ``Identifying chinese
  microblog users with high suicide probability using internet-based profile
  and linguistic features: classification model,'' \emph{JMIR mental health},
  vol.~2, no.~2, p. e17, 2015.

\bibitem{o2015detecting}
B.~O'Dea, S.~Wan, P.~J. Batterham, A.~L. Calear, C.~Paris, and H.~Christensen,
  ``Detecting suicidality on twitter,'' \emph{Internet Interventions}, vol.~2,
  no.~2, pp. 183--188, 2015.

\bibitem{coppersmith2015quantifying}
G.~Coppersmith, R.~Leary, E.~Whyne, and T.~Wood, ``Quantifying suicidal
  ideation via language usage on social media,'' in \emph{Joint Statistics
  Meetings Proceedings, Statistical Computing Section, JSM}, 2015.

\bibitem{parraga2019unsupervised}
J.~Parraga-Alava, R.~A. Caicedo, J.~M. G{\'o}mez, and M.~Inostroza-Ponta, ``An
  unsupervised learning approach for automatically to categorize potential
  suicide messages in social media,'' in \emph{2019 38th International
  Conference of the Chilean Computer Science Society}.\hskip 1em plus 0.5em
  minus 0.4em\relax IEEE, 2019, pp. 1--8.

\bibitem{fodeh2019using}
S.~Fodeh, T.~Li, K.~Menczynski, T.~Burgette, A.~Harris, G.~Ilita, S.~Rao,
  J.~Gemmell, and D.~Raicu, ``Using machine learning algorithms to detect
  suicide risk factors on twitter,'' in \emph{2019 International Conference on
  Data Mining Workshops (ICDMW)}.\hskip 1em plus 0.5em minus 0.4em\relax IEEE,
  2019, pp. 941--948.

\bibitem{ji2020suicidal}
S.~Ji, X.~Li, Z.~Huang, and E.~Cambria, ``Suicidal ideation and mental disorder
  detection with attentive relation networks,'' \emph{arXiv preprint
  arXiv:2004.07601}, 2020.

\bibitem{DBLP:journals/phat/MalhotraJ20}
\BIBentryALTinterwordspacing
A.~Malhotra and R.~Jindal, ``Multimodal deep learning based framework for
  detecting depression and suicidal behaviour by affective analysis of social
  media posts,'' \emph{{EAI} Endorsed Trans. Pervasive Health Technol.},
  vol.~6, no.~21, p.~e1, 2020. [Online]. Available:
  \url{https://doi.org/10.4108/eai.13-7-2018.164259}
\BIBentrySTDinterwordspacing

\bibitem{DBLP:journals/ijkbo/Bouarara20}
\BIBentryALTinterwordspacing
H.~A. Bouarara, ``Detection and prevention of twitter users with suicidal
  self-harm behavior,'' \emph{{IJKBO}}, vol.~10, no.~1, pp. 49--61, 2020.
  [Online]. Available: \url{https://doi.org/10.4018/IJKBO.2020010103}
\BIBentrySTDinterwordspacing

\bibitem{Sawhney2018}
R.~Sawhney, P.~Manchanda, R.~Singh, and S.~Aggarwal, ``A computational approach
  to feature extraction for identification of suicidal ideation in tweets,'' in
  \emph{Proceedings of the ACL Student Research Workshop}, 2018, pp. 91--98.

\bibitem{lei2019}
L.~Cao, H.~Zhang, L.~Feng, Z.~Wei, X.~Wang, N.~Li, and X.~He, ``Latent suicide
  risk detection on microblog via suicide-oriented word embeddings and layered
  attention,'' in \emph{Proceedings of the 2019 conference on empirical methods
  in natural language processing}, 2019.

\bibitem{Pradyumna2019b}
R.~Mishra, P.~P. Sinha, R.~Sawhney, D.~Mahata, P.~Mathur, and R.~R. Shah,
  ``Snap-batnet: Cascading author profiling and social network graphs for
  suicide ideation detection on social media,'' in \emph{Proceedings of the
  2019 Conference of the North American Chapter of the Association for
  Computational Linguistics: Student Research Workshop}, 2019, pp. 147--156.

\bibitem{cooperman2005suicidal}
N.~A. Cooperman and J.~M. Simoni, ``Suicidal ideation and attempted suicide
  among women living with hiv/aids,'' \emph{Journal of behavioral medicine},
  vol.~28, no.~2, pp. 149--156, 2005.

\bibitem{self-affirmation1988}
C.~Steele, ``The psychology of self-affirmation: sustaining the integrity of
  the self,'' \emph{Advances in experimental social psychology}, vol.~21,
  no.~2, pp. 261--302, 1988.

\bibitem{Rich1987}
A.~R. Rich and R.~L. Bonner, ``Concurrent validity of a stress-vulnerability
  model of suicidal ideation and behavior: A follow-up study,'' \emph{Suicide
  and Life - Threatening Behavior}, vol.~17, no.~4, pp. 265--270, 1987.

\bibitem{bluml2013personality}
V.~Bl{\"u}ml, N.~D. Kapusta, S.~Doering, E.~Br{\"a}hler, B.~Wagner, and
  A.~Kersting, ``Personality factors and suicide risk in a representative
  sample of the german general population,'' \emph{PlOS one}, vol.~8, no.~10,
  p. e76646, 2013.

\bibitem{greenspon2014there}
T.~S. Greenspon, ``Is there an antidote to perfectionism?'' \emph{Psychology in
  the Schools}, vol.~51, no.~9, pp. 986--998, 2014.

\bibitem{ML15}
M.~Lv, A.~Li, T.~Liu, and T.~Zhu, ``Creating a chinese suicide dictionary for
  identifying suicide risk on social media,'' \emph{PeerJ}, p.
  https://peerj.com/articles/1455/, 2015.

\bibitem{LM14}
L.~Mandelli, F.~Nearchou, C.~Vaiopoulos, C.~Stefanis, S.~Vitoratou,
  A.~Serretti, and N.~Stefanis, ``Neuroticism, social network, stressful life
  events: Association with mood disorders, depressive symptoms and suicidal
  ideation in a community sample of women,'' \emph{J. of Psychiatry Research},
  pp. 38--44, 2014.

\bibitem{LiQiJournal}
Q.~Li, Y.~Xue, L.~Zhao, J.~Jia, and L.~Feng, ``Analyzing and identifying teens
  stressful periods and stressor events from a microblog,'' \emph{IEEE Journal
  of Biomedical and Health Informatics (J-BHI)}, vol.~21, no.~5, pp. 1--15,
  2017.

\bibitem{coppersmith2014quantifying}
G.~Coppersmith, M.~Dredze, and C.~Harman, ``Quantifying mental health signals
  in twitter,'' in \emph{Proceedings of the workshop on computational
  linguistics and clinical psychology: From linguistic signal to clinical
  reality}, 2014, pp. 51--60.

\bibitem{xiao2018bertservice}
H.~Xiao, ``bert-as-service,'' \url{https://github.com/hanxiao/bert-as-service},
  2018.

\bibitem{he2016deep}
K.~He, X.~Zhang, S.~Ren, and J.~Sun, ``Deep residual learning for image
  recognition,'' in \emph{Proceedings of the IEEE conference on computer vision
  and pattern recognition}, 2016, pp. 770--778.

\bibitem{du2020adversarial}
C.~Du, H.~Sun, J.~Wang, Q.~Qi, and J.~Liao, ``Adversarial and domain-aware bert
  for cross-domain sentiment analysis,'' in \emph{Proceedings of the 58th
  Annual Meeting of the Association for Computational Linguistics}, 2020, pp.
  4019--4028.

\bibitem{chen2020distilling}
Y.-C. Chen, Z.~Gan, Y.~Cheng, J.~Liu, and J.~Liu, ``Distilling knowledge
  learned in bert for text generation,'' in \emph{Proceedings of the 58th
  Annual Meeting of the Association for Computational Linguistics}, 2020, pp.
  7893--7905.

\bibitem{zha2020adversarial}
Z.-J. Zha, J.~Liu, D.~Chen, and F.~Wu, ``Adversarial attribute-text embedding
  for person search with natural language query,'' \emph{IEEE Transactions on
  Multimedia}, 2020.

\bibitem{dai2019deep}
P.~Dai, H.~Zhang, and X.~Cao, ``Deep multi-scale context aware feature
  aggregation for curved scene text detection,'' \emph{IEEE Transactions on
  Multimedia}, vol.~22, no.~8, pp. 1969--1984, 2019.

\bibitem{JS69}
J.~Sabbath, ``The suicidal adolescent: The expendable child,'' \emph{J. of the
  American Academy of Child Psychiatry}, pp. 272--285, 1969.

\bibitem{lv2015creating}
M.~Lv, A.~Li, T.~Liu, and T.~Zhu, ``Creating a chinese suicide dictionary for
  identifying suicide risk on social media,'' \emph{PeerJ}, vol.~3, p. e1455,
  2015.

\bibitem{huang2012development}
C.-L. Huang, C.~K. Chung, N.~Hui, Y.-C. Lin, Y.-T. Seih, B.~C. Lam, W.-C. Chen,
  M.~H. Bond, and J.~W. Pennebaker, ``The development of the chinese linguistic
  inquiry and word count dictionary.'' \emph{Chinese Journal of Psychology},
  2012.

\bibitem{kipf2016semi}
\BIBentryALTinterwordspacing
T.~N. Kipf and M.~Welling, ``Semi-supervised classification with graph
  convolutional networks,'' in \emph{5th International Conference on Learning
  Representations, {ICLR} 2017, Toulon, France, April 24-26, 2017, Conference
  Track Proceedings}.\hskip 1em plus 0.5em minus 0.4em\relax OpenReview.net,
  2017. [Online]. Available: \url{https://openreview.net/forum?id=SJU4ayYgl}
\BIBentrySTDinterwordspacing

\bibitem{hamilton2017inductive}
\BIBentryALTinterwordspacing
W.~L. Hamilton, Z.~Ying, and J.~Leskovec, ``Inductive representation learning
  on large graphs,'' in \emph{Advances in Neural Information Processing Systems
  2017, December 4-9, 2017, Long Beach, CA, {USA}}, 2017, pp. 1024--1034.
  [Online]. Available:
  \url{http://papers.nips.cc/paper/6703-inductive-representation-learning-on-large-graphs}
\BIBentrySTDinterwordspacing

\bibitem{ParentingStyle1993}
N.~Darling and L.~Steinberg, ``Parenting style as context: an integrative
  model,'' \emph{Psychological Bulletin}, vol. 113, no.~3, pp. 487--496, 1993.

\bibitem{ParentingStyle2001}
K.~Lai and C.~Mcbride-Chang, ``Suicidal ideation, parenting style, and family
  climate among hong kong adolescents,'' \emph{Int J. Psychology}, vol.~36,
  no.~2, pp. 81--87, 2001.

\end{thebibliography}
